\documentclass{article}

\usepackage{microtype}
\usepackage{graphicx}
\usepackage{booktabs} 
\usepackage{times}
\usepackage{hyperref}

\usepackage{amsmath,amsfonts,bm}

\newcommand{\cB}{\mathbf{c}}

\newcommand{\xB}{\mathbf{x}}
\newcommand{\XB}{\mathbf{X}}

\def\eqref#1{equation~\ref{#1}}

\def\1{\bm{1}}

\DeclareMathAlphabet{\mathsfit}{\encodingdefault}{\sfdefault}{m}{sl}
\SetMathAlphabet{\mathsfit}{bold}{\encodingdefault}{\sfdefault}{bx}{n}

\usepackage[dvipsnames]{xcolor}
\usepackage{hyperref}
\usepackage{url}
\usepackage{booktabs}      
\usepackage{amsfonts}     
\usepackage{nicefrac} 
\usepackage{microtype}  
\usepackage{natbib}

\usepackage{graphicx}
\usepackage{algorithm}
\usepackage[noend]{algpseudocode}
\usepackage{caption}
\usepackage{subcaption}
\usepackage{multirow}
\usepackage{makecell}
\usepackage{placeins}
\usepackage[export]{adjustbox} 
\usepackage{amsmath}
\usepackage{amsthm}
\usepackage{verbatim}
\usepackage{float}
\usepackage{wrapfig}
\usepackage{subcaption}
\usepackage{dsfont}

\newcommand{\maybebf}[0]{}

\newcommand{\matr}[1]{\mathbf{#1}}

\usepackage{xr}
\makeatletter
\newcommand*{\addFileDependency}[1]{
  \typeout{(#1)}
  \@addtofilelist{#1}
  \IfFileExists{#1}{}{\typeout{No file #1.}}
}
\makeatother

\usepackage[accepted]{icml2021}

\icmltitlerunning{Neural Architecture Search without Training}

\begin{document}

\twocolumn[
\icmltitle{Neural Architecture Search without Training}

\icmlsetsymbol{equal}{*}

\begin{icmlauthorlist}
\icmlauthor{Joseph Mellor}{ush}
\icmlauthor{Jack Turner}{inf}
\icmlauthor{Amos Storkey}{inf}
\icmlauthor{Elliot J. Crowley}{eng}
\end{icmlauthorlist}

\icmlaffiliation{ush}{Usher Institute, University of Edinburgh}
\icmlaffiliation{inf}{School of Informatics, University of Edinburgh}
\icmlaffiliation{eng}{School of Engineering, University of Edinburgh}

\icmlcorrespondingauthor{Joseph Mellor}{joe.mellor@ed.ac.uk}

\icmlkeywords{Machine Learning, ICML}

\vskip 0.3in
]

\printAffiliationsAndNotice{}

\begin{abstract}
The time and effort involved in hand-designing deep neural networks is immense. This has prompted the development of Neural Architecture Search (NAS) techniques to automate this design. However, NAS algorithms tend to be slow and expensive; they need to train vast numbers of candidate networks to inform the search process. This could be alleviated if we could partially predict a network's trained accuracy from its initial state. In this work, we examine the overlap of activations between datapoints in {\it untrained} networks and motivate how this can give a measure which is usefully indicative of a network’s {\it trained} performance. We incorporate this measure into a simple algorithm that allows us to search for powerful networks without any training in a matter of seconds on a single GPU, and verify its effectiveness on NAS-Bench-101, NAS-Bench-201, NATS-Bench, and Network Design Spaces. Our approach can be readily combined with more expensive search methods; we examine a simple adaptation of regularised evolutionary search. Code for reproducing our experiments is available at~\url{https://github.com/BayesWatch/nas-without-training}.
\end{abstract}

\section{Introduction}
The success of deep learning in computer vision is in no small part due to the insight and engineering efforts of human experts, allowing for the creation of powerful architectures for widespread adoption~\citep{krizhevsky2012imagnet,simonyan2015very,he2016deep,szegedy2016rethinking,huang2017densely}. However, this manual design is costly, and becomes increasingly more difficult as networks get larger and more complicated. Because of these challenges, the neural network community has seen a shift from designing architectures to designing algorithms that {\it search} for candidate architectures~\citep{elsken2019neural,wistuba2019survey}. These Neural Architecture Search (NAS) algorithms are capable of automating the discovery of effective architectures~\citep{zoph2017neural,zoph2018learning,pham2018efficient,tan2019mnasnet,liu2019darts,real2019regularized}.

\begin{figure*}[!h]

\centering
    \begin{subfigure}{.49\textwidth}
    \includegraphics[width=.99\textwidth]{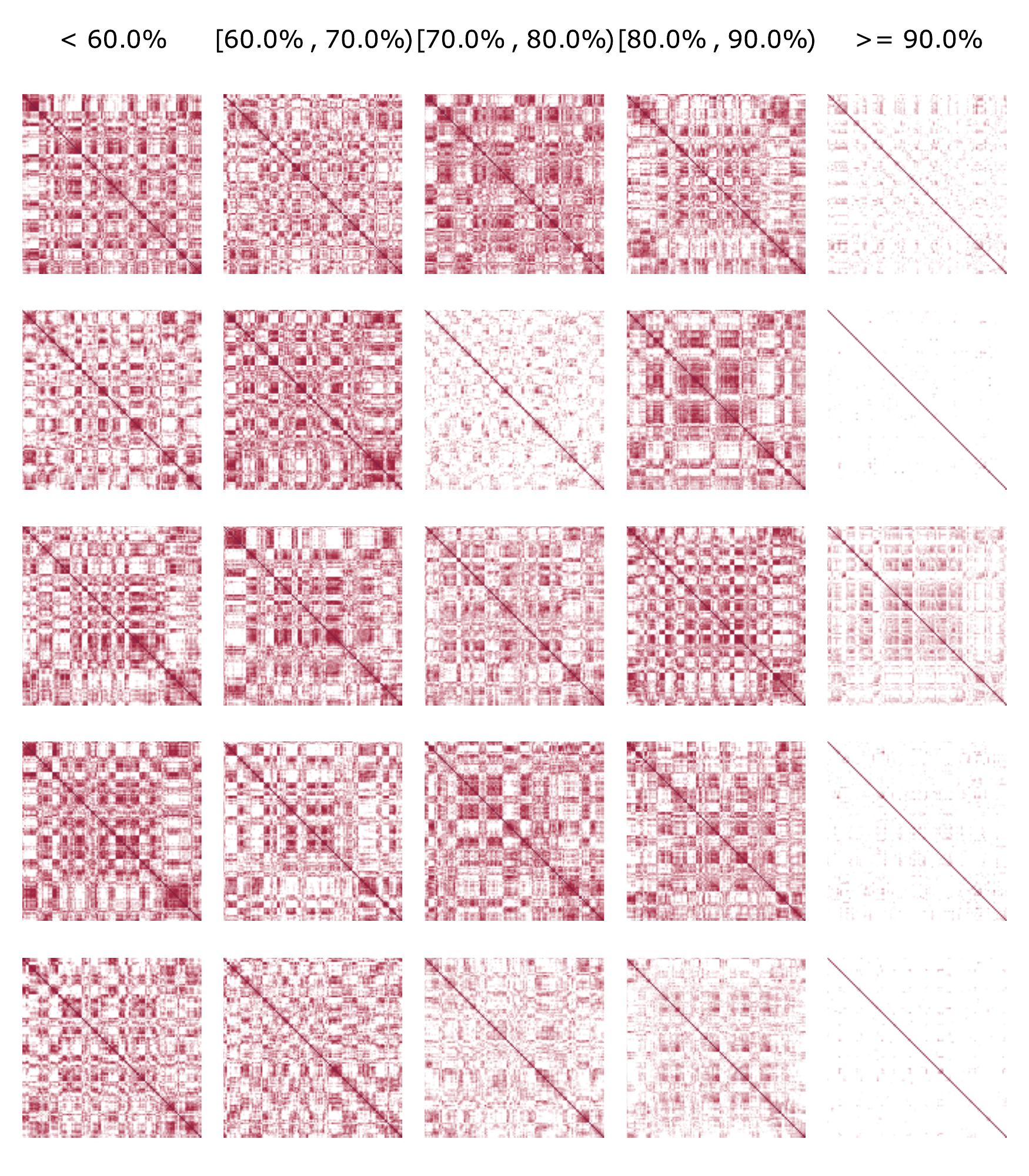}
    \caption{NAS-Bench-201}
    \label{fig:activationhamming201}
    \end{subfigure}~
    \begin{subfigure}{.49\textwidth}
    \includegraphics[width=.99\textwidth]{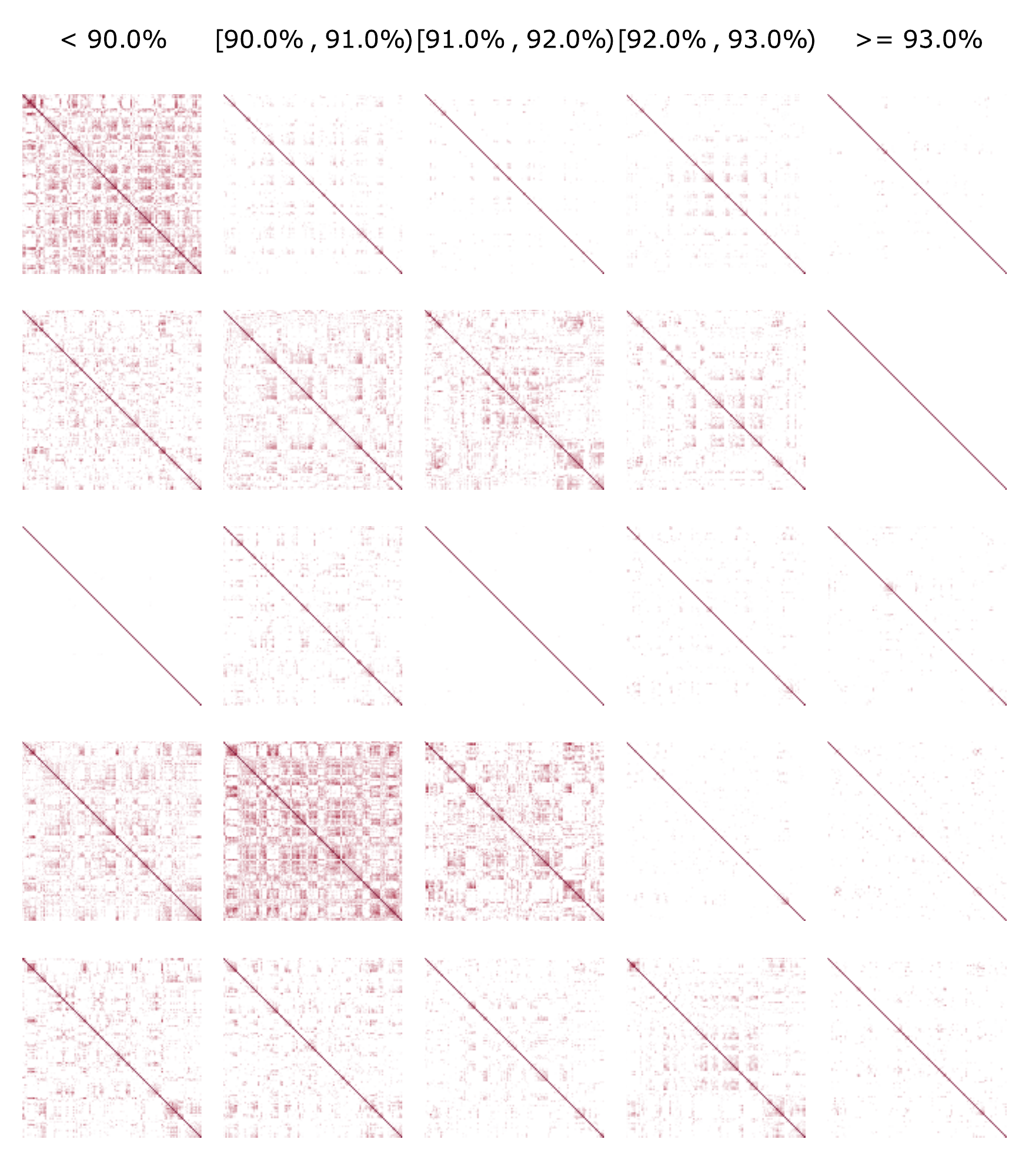}
    \caption{NDS-DARTS}
    \label{fig:activationhammingdarts}
    \end{subfigure}

\caption{{\color{black} $\matr{K}_H$ for a mini-batch of 128 CIFAR-10 images for {\maybebf untrained} architectures in (a) NAS-Bench-201~\citep{Dong2020NAS-Bench-201} and (b) NDS-DARTS~\citep{radosavovic2019network}. $\matr{K}_H$ in these plots is normalised so that the diagonal entries are $1$. The $\matr{K}_H$ are sorted into columns based on the final CIFAR-10 validation accuracy {\maybebf when trained}. Darker regions have higher similarity. The profiles are distinctive; the $\matr{K}_H$ for good architectures in both search spaces have less similarity between different images. We can use $\matr{K}_H$ for an untrained network to predict its final performance without any training. Note that (b) covers a tighter accuracy range than (a), which may explain it being less distinctive.}}
    \label{fig:normhamming} 
\end{figure*}

NAS algorithms are broadly based on the seminal work of~\cite{zoph2017neural}. A controller network generates an architecture proposal, which is then trained to provide a signal to the controller through REINFORCE~\citep{williams1992simple}, which then produces a new proposal, and so on. Training a network for every controller update is extremely expensive; utilising 800 GPUs for 28 days in~\cite{zoph2017neural}. Subsequent work has sought to ameliorate this by (i) learning stackable cells instead of whole networks~\citep{zoph2018learning} and (ii) incorporating {\it weight sharing}; allowing candidate networks to share weights to allow for joint training~\citep{pham2018efficient}. These contributions have accelerated the speed of NAS algorithms e.g.\ to half a day on a single GPU in~\cite{pham2018efficient}.

For some practitioners, NAS is still too slow; being able to perform NAS quickly (i.e.\ in seconds) would be immensely useful in the hardware-aware setting where a separate search is typically required for each device and task~\citep{wu2019fbnet,tan2019mnasnet}. This could be achieved if NAS could be performed {\it without any network training}. In this paper {\maybebf we show that this is possible}. We explore NAS-Bench-101~\citep{ying2019bench}, NAS-Bench-201~\citep{Dong2020NAS-Bench-201}, NATS-Bench~\citep{dong2021nats}, and Network Design Spaces (NDS,~\citealp{radosavovic2019network}), and examine the overlap of activations between datapoints in a mini-batch for an {\maybebf untrained} network (Section~\ref{sec:scoring}). The linear maps of the network are uniquely identified by a binary code corresponding to the activation pattern of the rectified linear units. The Hamming distance between these binary codes can be used to define a kernel matrix (which we denote by $\matr{K}_H$) which is distinctive for networks that perform well; this is immediately apparent from visualisation alone across two distinct search spaces (Figure~\ref{fig:normhamming}). We devise a score based on $\matr{K}_H$ and perform an ablation study to demonstrate its robustness to inputs and network initialisation.

We incorporate our score into a simple search algorithm that {\maybebf doesn't require training} (Section~\ref{sec:nas}). This allows us to perform architecture search quickly, for example, on CIFAR-10~\citep{krizhevsky2009learning} we are able to search for a network that achieves $92.81\%$ accuracy in 30 seconds within the NAS-Bench-201 search space; several orders of magnitude faster than traditional NAS methods for a modest change in final accuracy. We also show how we can combine our approach with regularised evolutionary search (REA,~\citealp{pham2018efficient}) as an example of how it can be readily integrated into existing NAS techniques.

We believe that this work is an important proof-of-concept for NAS without training. The large resource and time costs associated with NAS can be avoided; our search algorithm uses a single GPU and is extremely fast. The benefit is two-fold, as we also show that we can integrate our approach into existing NAS techniques for scenarios where obtaining as high an accuracy as possible is of the essence.

\begin{figure*}[!h]
\centering

    \includegraphics[width=.99\textwidth]{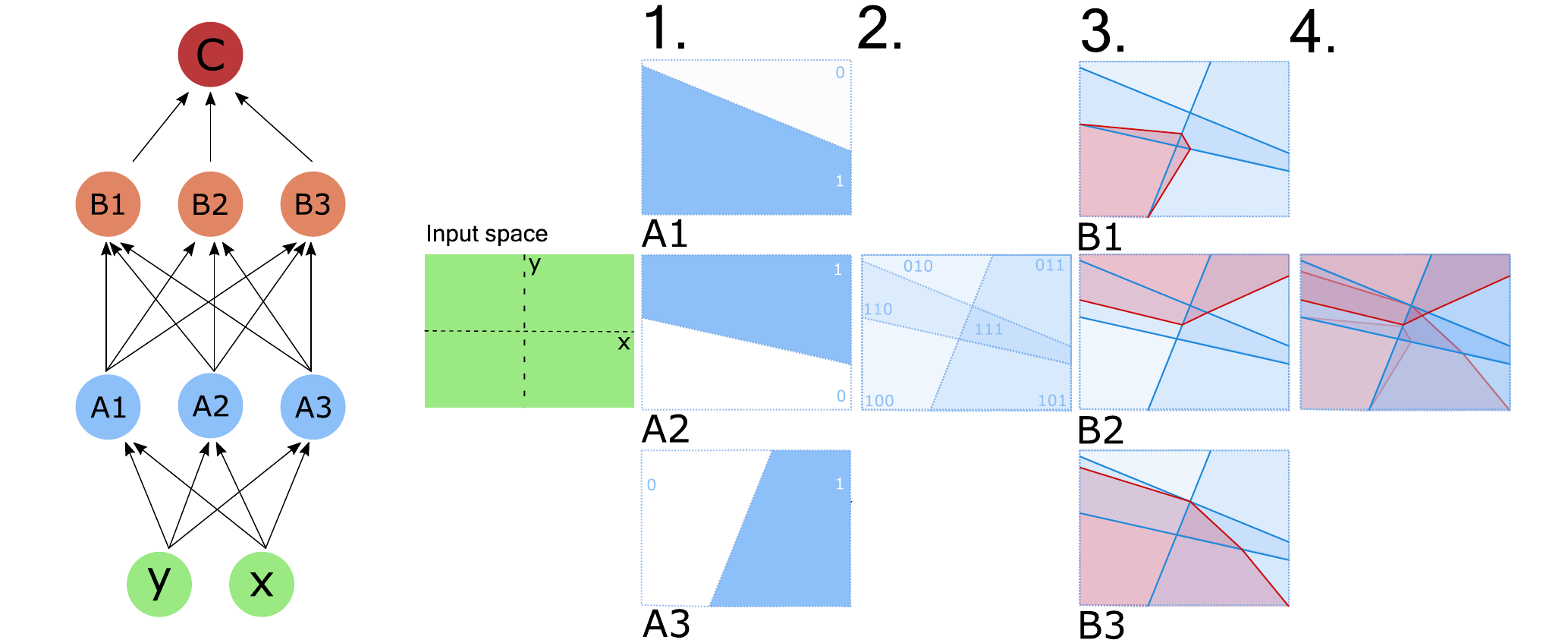}
\caption{Visualising how binary activation codes of ReLU units correspond to linear regions.
{\bf 1:} Each ReLU node A$i$ splits the input into an active ($>0$) and inactive region We label the active region 1 and inactive 0. {\bf 2:} The active/inactive regions associated with each node A$i$ intersect. Areas of the input space with the same activation pattern are co-linear. Here we show the intersection of the A nodes and give the code for the linear regions. Bit $i$ of the code corresponds to whether node A$i$ is active. {\bf 3:} The ReLU nodes B of the next layer divides the space further into active and inactive regions. {\bf 4:} Each linear region at a given node can be uniquely defined by the activation pattern of all the ReLU nodes that preceded it.}
    \label{fig:locallinearcodes} 
\end{figure*}

\section{Background}
\label{sec:lit}

Designing a neural architecture by hand is a challenging and time-consuming task. It is extremely difficult to intuit where to place connections, or which operations to use. This has prompted an abundance of research into neural architecture search (NAS); the automation of the network design process. In the pioneering work of~\cite{zoph2017neural}, the authors use an RNN controller to generate descriptions of candidate networks. Candidate networks are trained, and used to update the controller using reinforcement learning to improve the quality of the candidates it generates. This algorithm is very expensive: searching for an architecture to classify CIFAR-10 required running 800 GPUs for 28 days. It is also inflexible; the final network obtained is fixed and cannot be scaled e.g.\ for use on mobile devices or for other datasets.

The subsequent work of~\cite{zoph2018learning} deals with these limitations. Inspired by the modular nature of successful hand-designed networks~\citep{simonyan2015very,he2016deep,huang2017densely}, they propose searching over neural building blocks, instead of over whole architectures. These building blocks, or {\it cells}, form part of a fixed overall network structure. Specifically, the authors search for a standard cell, and a reduced cell (incorporating pooling) for CIFAR-10 classification. These are then used as the building blocks of a larger network for ImageNet~\citep{russakovsky2015imagenet} classification. While more flexible---the number of cells can be adjusted according to budget---and cheaper, owing to a smaller search space, this technique still utilised 500 GPUs across 4 days.

ENAS~\citep{pham2018efficient} reduces the computational cost of searching by allowing multiple candidate architectures to share weights. This facilitates the simultaneous training of candidates, reducing the search time on CIFAR-10 to half a day on a single GPU. Weight sharing has seen widespread adoption in a host of NAS algorithms~\citep{liu2019darts,luo2018neural,cai2019proxylessnas,xie2019snas,brock2018smash}. However, there is evidence that it inhibits the search for optimal architectures~\citep{yu2020evaluating}, exposing random search as an extremely effective NAS baseline~\citep{yu2020evaluating,li2019random}. There also remains the problem that the search spaces are still so vast---there are $1.6\times 10^{29}$ possible architectures in~\cite{pham2018efficient} for example---that it is impossible to identify the best networks and  demonstrate that NAS algorithms find them.

An orthogonal direction for identifying good architectures is the estimation of accuracy prior to training~\citep{deng2017peephole,istrate2019tapas}, although these differ from this work in that they rely on training a predictive model, rather than investigating more fundamental architectural properties. 
Since its inception others have explored our work and the ideas therein in interesting directions. Of most interest from our perspective are \citet{abdelfattah2021zerocost} who integrate training-free heuristics into existing more-expensive search strategies to improve their performance  as we do in this paper. 
\citet{park2020towards} use the correspondence between wide neural networks and Gaussian processes to motivate using as a heuristic the validation accuracy of a Monte-Carlo approximated neural network Gaussian process conditioned on training data. \citet{chen2021neural} propose two further heuristics---one based on the condition number of the neural tangent kernel~\citep{jacot2018neural} at initialisation and the other based on the number of unique linear regions that partition training data at initialisation---with a proposed strategy to combine these heuristics into a stronger one.

\subsection{NAS Benchmarks}
\label{sec:nasbench101}
A major barrier to evaluating the effectiveness of a NAS algorithm is that the search space (the set of all possible networks) is too large for exhaustive evaluation. This has led to the creation of several benchmarks~\citep{ying2019bench,Zela2020NAS-Bench-1Shot1:,Dong2020NAS-Bench-201,dong2021nats} that consist of tractable NAS search spaces, and metadata for the training of networks within that search space. Concretely, this means that it is now possible to determine whether an algorithm is able to search for a good network. In this work we utilise NAS-Bench-101~\citep{ying2019bench}, NAS-Bench-201~\citep{Dong2020NAS-Bench-201}, and NATS-Bench~\citep{dong2021nats} to evaluate the effectiveness of our approach. NAS-Bench-101 consists of 423,624 neural networks that have been trained exhaustively, with three different initialisations, on the CIFAR-10 dataset for 108 epochs. NAS-Bench-201 consists of 15,625 networks trained multiple times on CIFAR-10, CIFAR-100, and ImageNet-16-120~\citep{chrabaszcz2017downsampled}. NATS-Bench~\citep{dong2021nats} comprises two search spaces: a topology search space (NATS-Bench TSS) which contains the same 15,625 networks as NAS-Bench 201; and a size search space (NATS-Bench SSS) which contains 32,768 networks where the number of channels for cells varies between these networks.
These benchmarks are described in detail in Appendix~\ref{appendix:nasbench}.

We also make use of the Network Design Spaces (NDS) dataset~\citep{radosavovic2019network}. Where the NAS benchmarks aim to compare search~\textit{algorithms}, NDS aims to compare the search~\textit{spaces} themselves. All networks in NDS use the DARTS~\citep{liu2019darts} skeleton. The networks are comprised of cells sampled from one of several NAS search spaces. Cells are sampled---and the resulting networks are trained---from each of AmoebaNet~\citep{real2019regularized}; DARTS~\citep{liu2019darts}; ENAS~\citep{pham2018efficient},  NASNet~\citep{zoph2017neural}, and
PNAS~\citep{liu2018progressive}. 

We denote each of these sets as NDS-AmoebaNet, NDS-DARTS, NDS-ENAS, NDS-NASNet, and NDS-PNAS respectively. Note that these sets contain networks of variable width and depth, whereas in e.g.\ the original DARTS search space these were fixed quantities.\footnote{NDS also contains a set of networks with fixed width and depth for the DARTS, ENAS, and PNAS cell search spaces. We provide experiments on these sets in Appendix~\ref{appendix:plots}.}

\section{Scoring Networks at Initialisation}
\label{sec:scoring}

Our goal is to devise a means to score a network architecture at initialisation in a way that is indicative of its final trained accuracy. This can either replace the expensive inner-loop training step in NAS, or better direct exploration in existing NAS algorithms.

Given a neural network with rectified linear units, we can, at each unit in each layer, identify a binary indicator as to whether the unit is inactive (the value is negative and hence is multiplied by zero) or active (in which case its value is multiplied by one). Fixing these indicator variables, it is well known that the network is now locally defined by a linear operator~\citep{hanin2019deep}; this operator is obtained by multiplying the linear maps at each layer interspersed with the binary rectification units. Consider a mini-batch of data $\XB = \{\xB_{i}\}_{i=1}^{N}$ mapped through a neural network as $f(\xB_i )$. The indicator variables from the rectified linear units in $f$ at $\xB_i$ form a binary code $\cB_i$ that defines the linear region.

The intuition to our approach is that the more similar the binary codes associated with two inputs are then the more challenging it is for the network to learn to separate these inputs. When two inputs have the same binary code, they lie within the same linear region of the network and so are particularly difficult to disentangle. Conversely, learning should prove easier when inputs are well separated. Figure~\ref{fig:locallinearcodes} visualises binary codes corresponding to linear regions. 

We use the Hamming distance $d_H(\cB_i, \cB_j)$ between two binary codes---induced by the untrained network at two inputs---as a measure of how dissimilar the two inputs are.

We can examine the correspondence between binary codes for the whole mini-batch by computing the kernel matrix

\begin{equation}
\matr{K}_H = 
\begin{pmatrix}

    N_A {-} d_H(\cB_1, \cB_1) & \cdots & N_A {-} d_H(\cB_1, \cB_N) \\
    \vdots & \ddots & \vdots \\
    N_A {-} d_H(\cB_N, \cB_1) & \cdots & N_A {-} d_H(\cB_N, \cB_N) \\
\end{pmatrix}
\label{eq:KH}
\end{equation}
where $N_A$ is the number of rectified linear units. 

We compute $\matr{K}_H$ for a random subset of NAS-Bench-201~\citep{Dong2020NAS-Bench-201} and NDS-DARTS \citep{radosavovic2019network} networks {\bf at initialisation} for a mini-batch of CIFAR-10 images. We plot normalised $\matr{K}_H$ for different trained accuracy bounds in Figure~\ref{fig:normhamming}.

\begin{figure*}[!ht]

     \centering
     \begin{subfigure}[b]{0.3\textwidth}
         \centering
         \includegraphics[width=\textwidth]{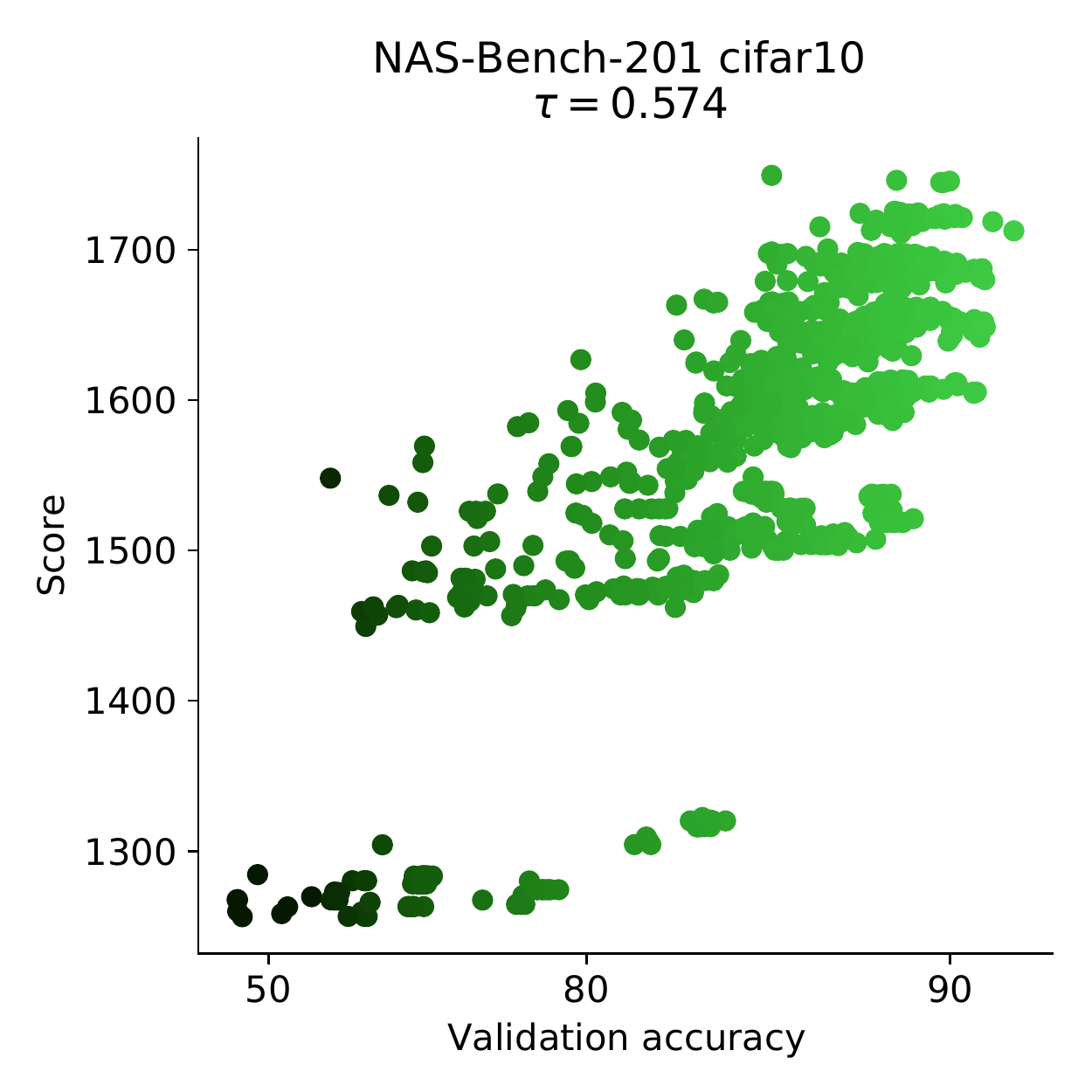}
         \caption{}
     \end{subfigure}
     \hfill
     \begin{subfigure}[b]{0.3\textwidth}
         \centering
         \includegraphics[width=\textwidth]{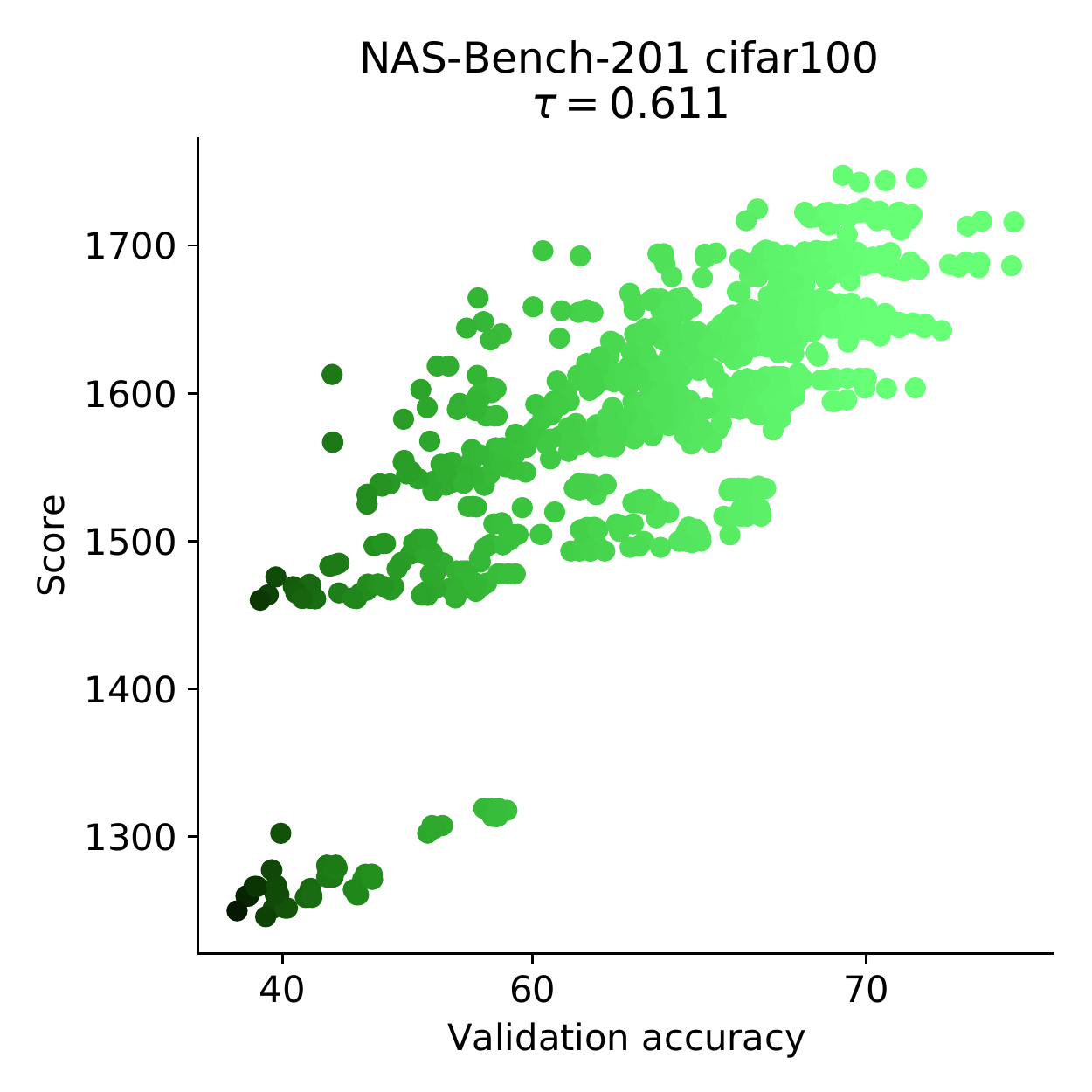}
         \caption{}
     \end{subfigure}
     \hfill
     \begin{subfigure}[b]{0.3\textwidth}
         \centering
         \includegraphics[width=\textwidth]{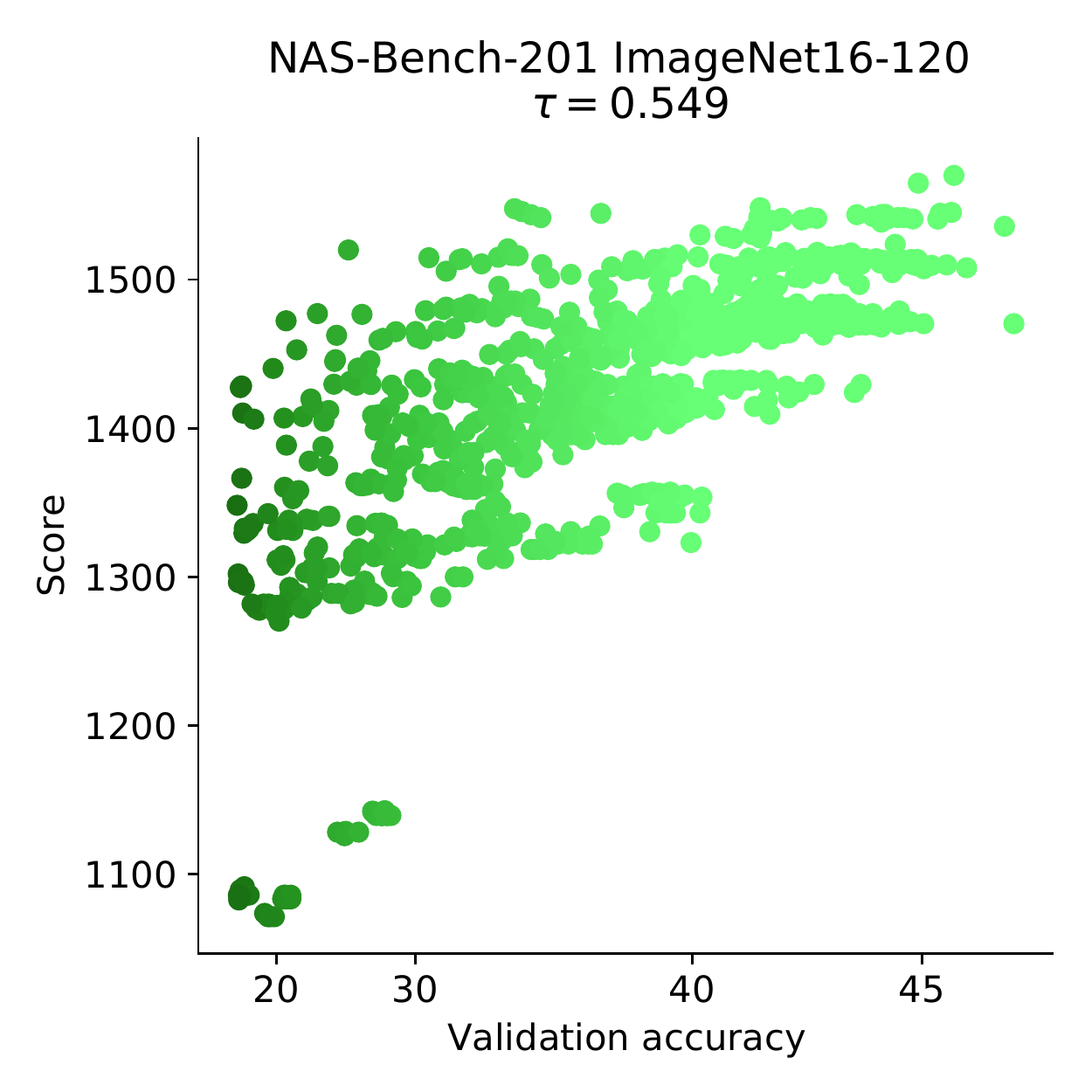}
         \caption{}
         \label{fig:pnas}
     \end{subfigure}

    \centering
     \begin{subfigure}[b]{0.3\textwidth}
         \centering

         \includegraphics[width=\textwidth]{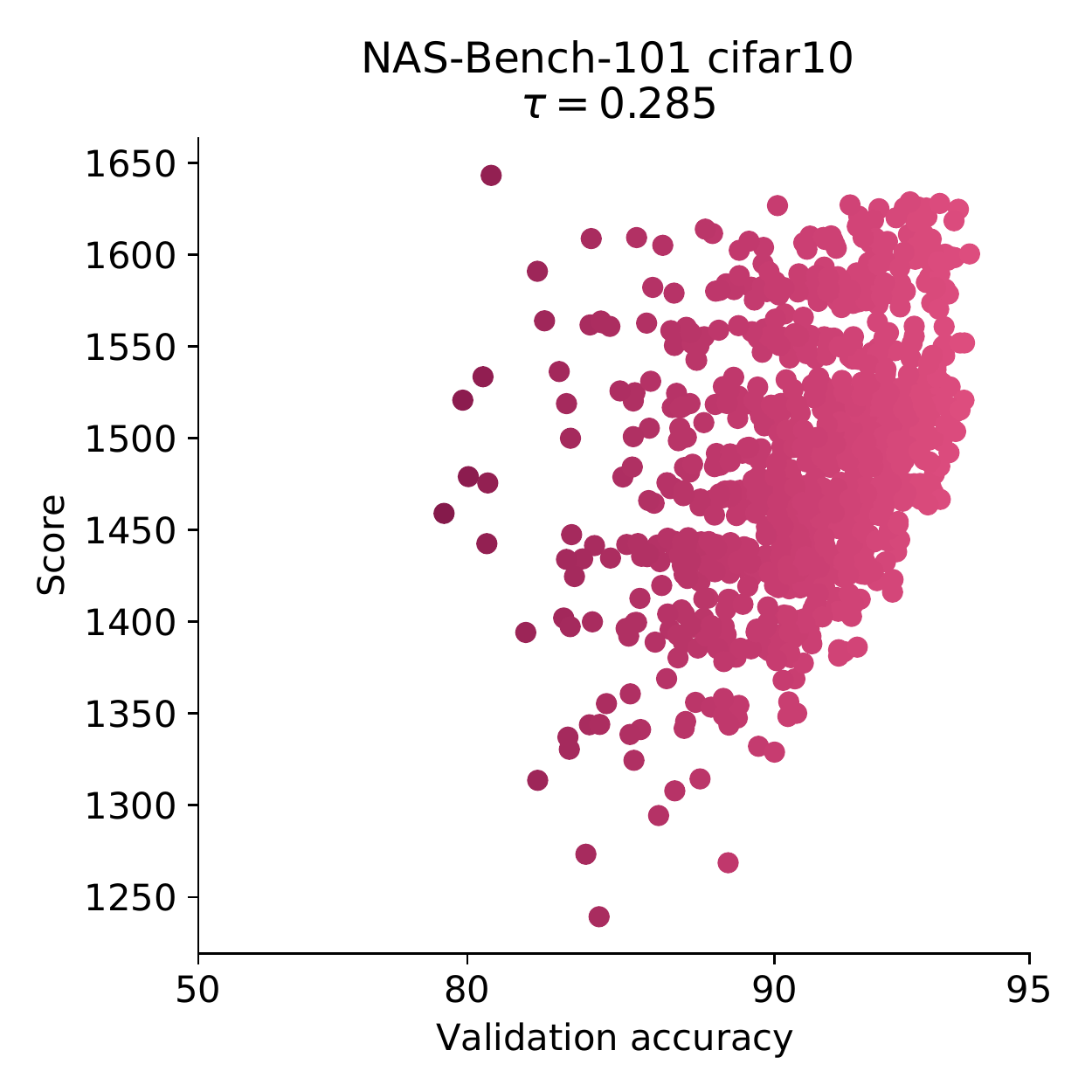}
         \caption{}
     \end{subfigure}
     \hfill
     \begin{subfigure}[b]{0.3\textwidth}
         \centering

         \includegraphics[width=\textwidth]{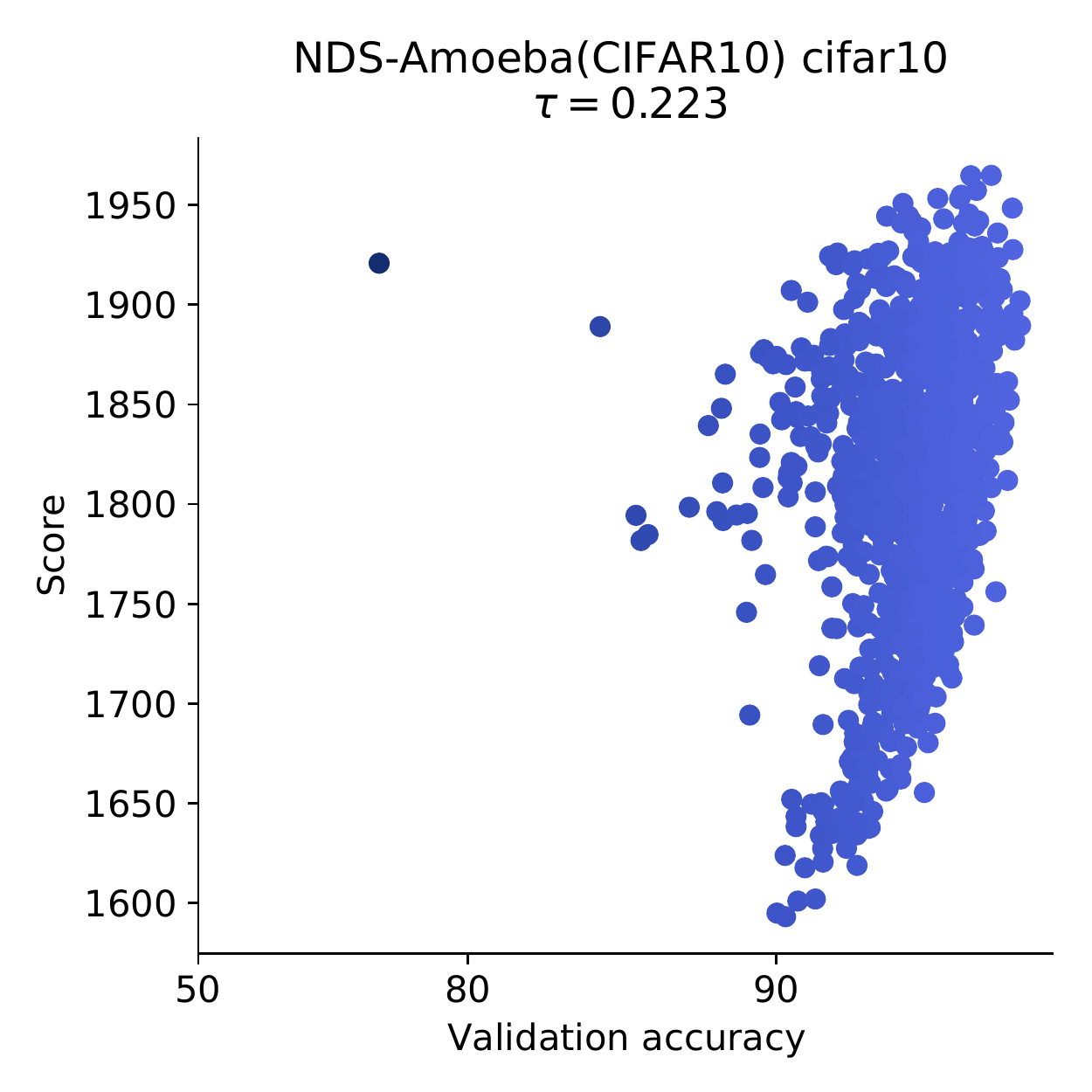}
         \caption{}
     \end{subfigure}
     \hfill
     \begin{subfigure}[b]{0.3\textwidth}
         \centering
         \includegraphics[width=\textwidth]{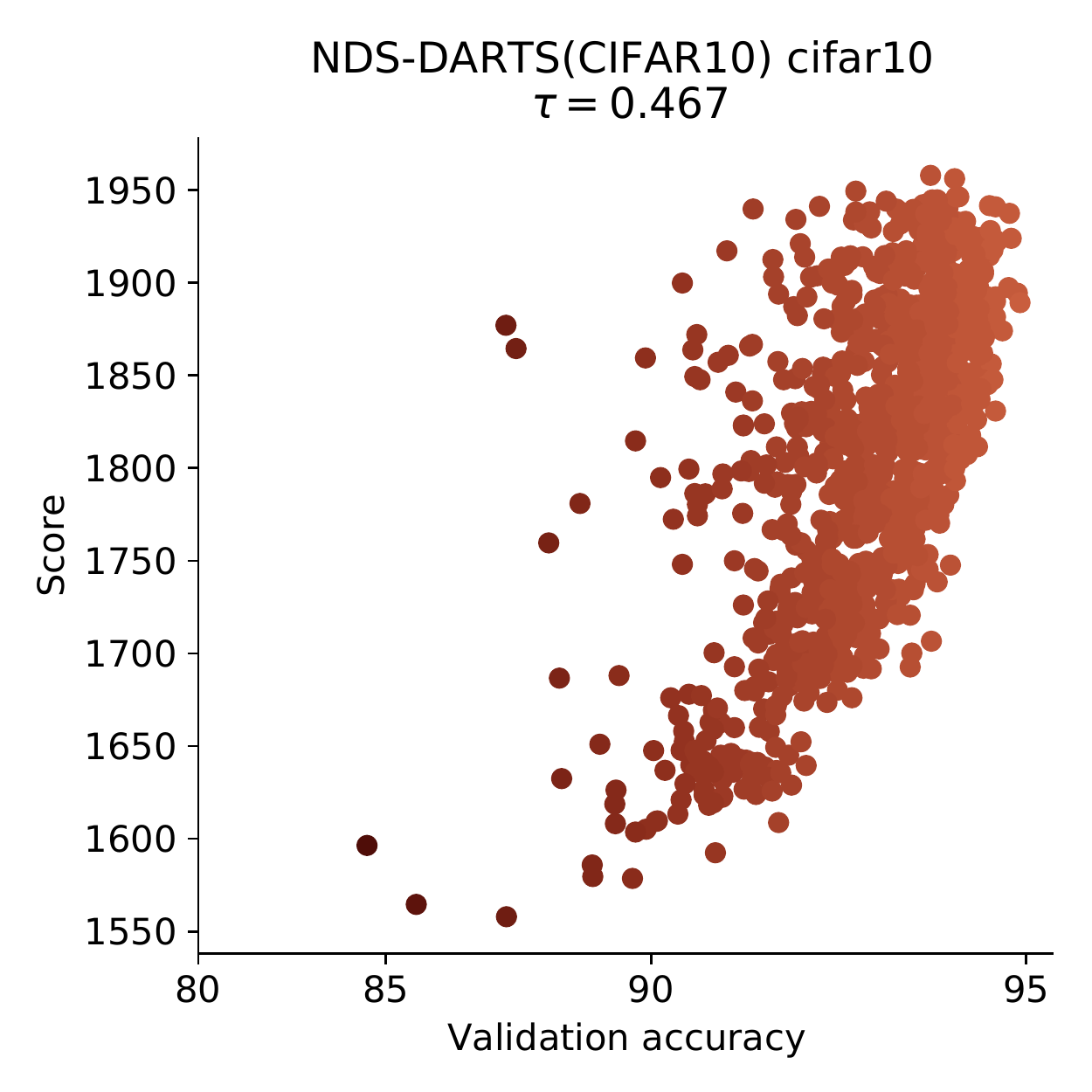}
         \caption{}
     \end{subfigure}

    \centering
     \begin{subfigure}[b]{0.3\textwidth}
         \centering

        \includegraphics[width=\textwidth]{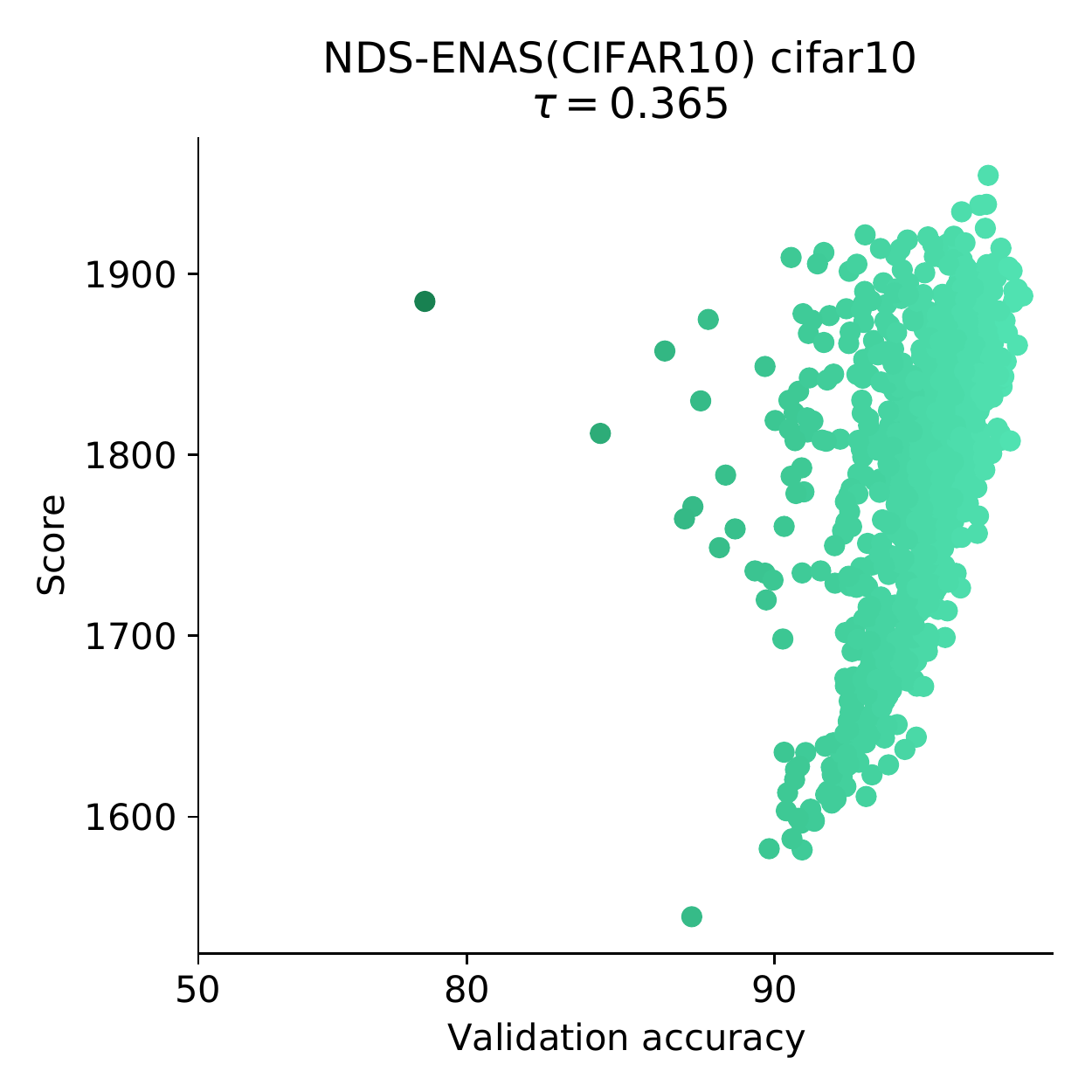}
         \caption{}
         \label{fig:amoeba}
     \end{subfigure}
     \hfill
          \begin{subfigure}[b]{0.3\textwidth}
         \centering

         \includegraphics[width=\textwidth]{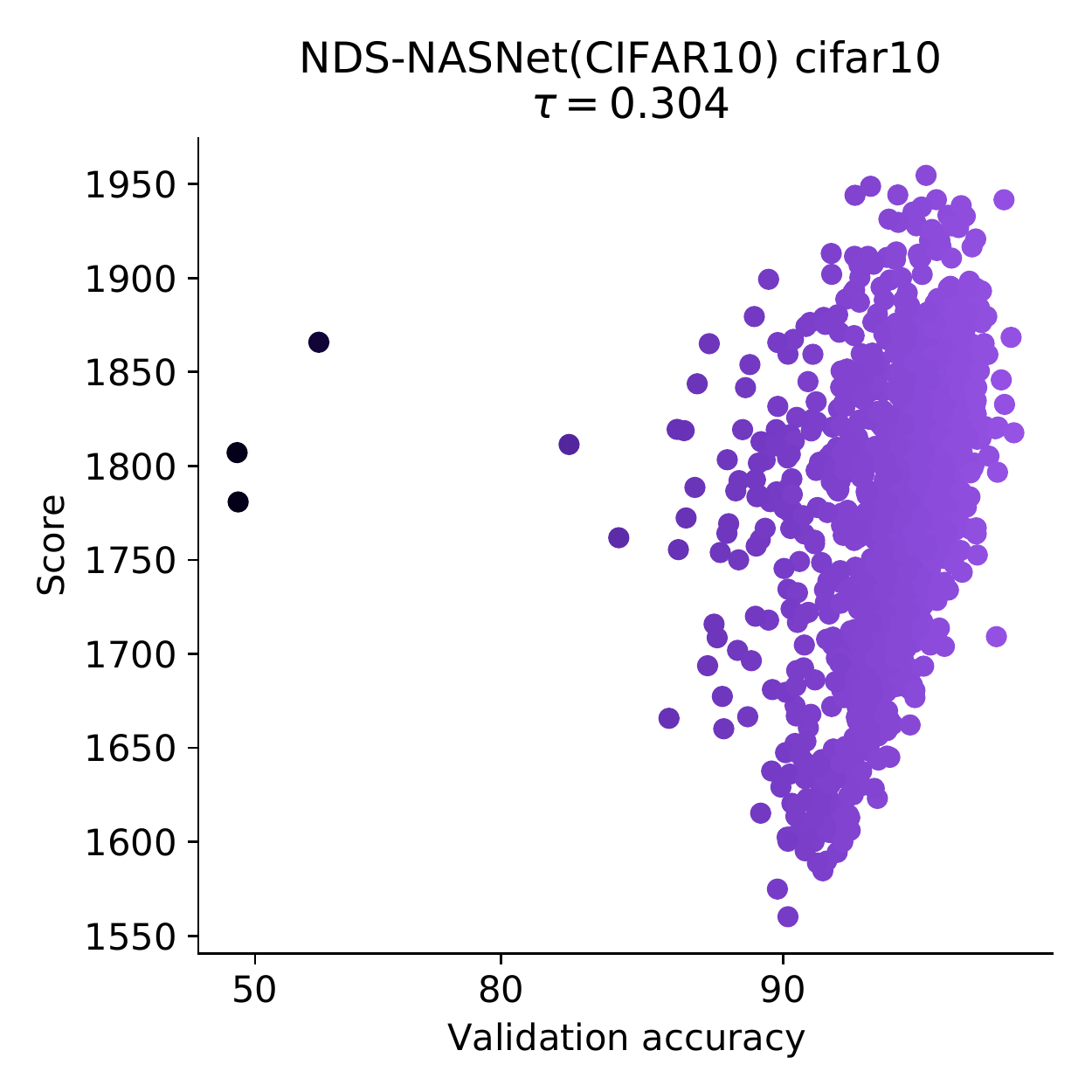}
         \caption{}
     \end{subfigure}
     \hfill
     \begin{subfigure}[b]{0.3\textwidth}
         \centering
         \includegraphics[width=\textwidth]{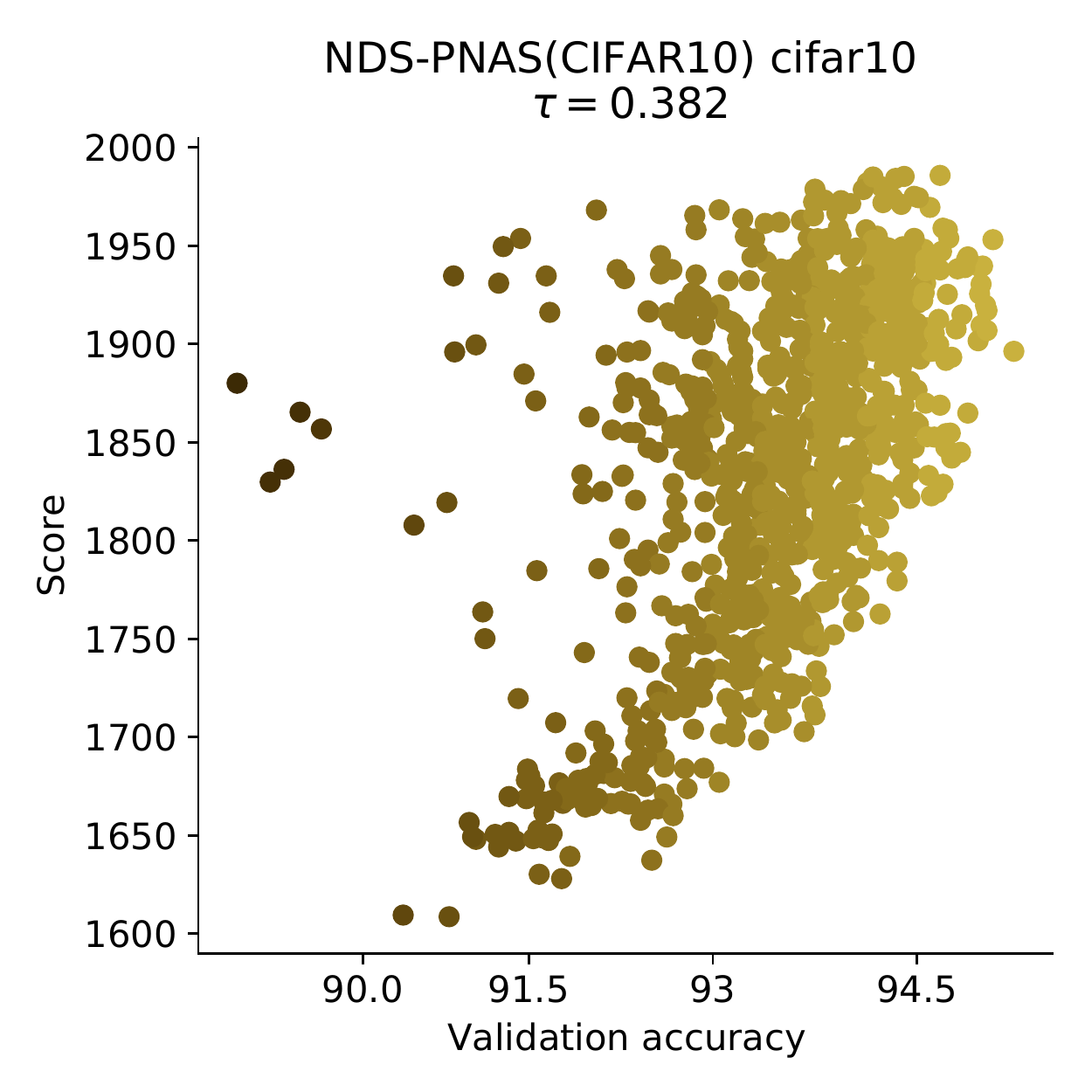}
         \caption{}
         \label{fig:resnet}
     \end{subfigure}

         \begin{tabular}{l p{.66\linewidth}}
             \raisebox{-.75\height}{\includegraphics[width=0.3\linewidth]{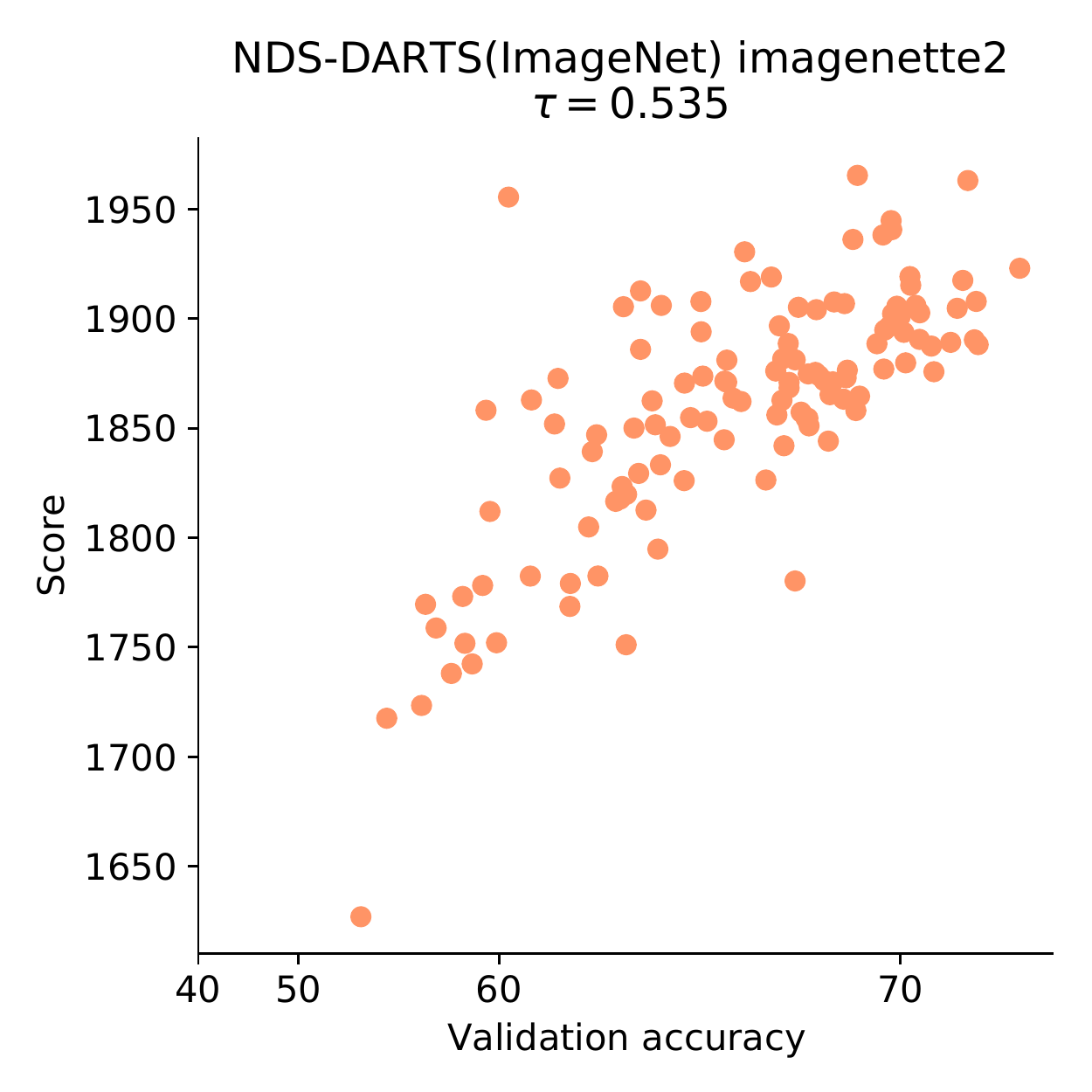}}
             & 
             \textit{Figure 3.} (a)-(i): Plots of our score for randomly sampled {\bf untrained} architectures in 
   NAS-Bench-201,
   NAS-Bench-101,
   NDS-Amoeba,
   NDS-DARTS,
   NDS-ENAS,
   NDS-NASNet,
   NDS-PNAS
   against validation accuracy when trained. The inputs when computing the score and the validation accuracy for each plot are from CIFAR-10 except for (b) and (c) which use CIFAR-100 and ImageNet16-120 respectively. (j): We include a plot from NDS-DARTS on ImageNet (121 networks provided) to illustrate that the score extends to more challenging datasets. We use a mini-batch from ImageNette2 which is a strict subset of ImageNet with only 10 classes. In all cases there is a noticeable correlation between the score for an untrained network and the final accuracy when trained.    \\
   \quad\quad\quad\quad\quad\quad\quad(j) & \\
         \end{tabular}

    \label{fig:scoreplots} 

\end{figure*}

These normalised kernel plots are very distinct; high performing networks have fewer off-diagonal elements with high similarity. We can use this observation to predict the final performance of untrained networks, in place of the expensive training step in NAS.
Specifically, we score networks using:

\begin{equation}
s = \log \vert \matr{K}_H \vert
\label{eq:score}
\end{equation}

Given two kernels with the same trace,~$s$ is higher for the kernel closest to diagonal. A higher score at initialisation implies improved final accuracy after training.

\begin{figure*}[!h]

\includegraphics[width=\linewidth]{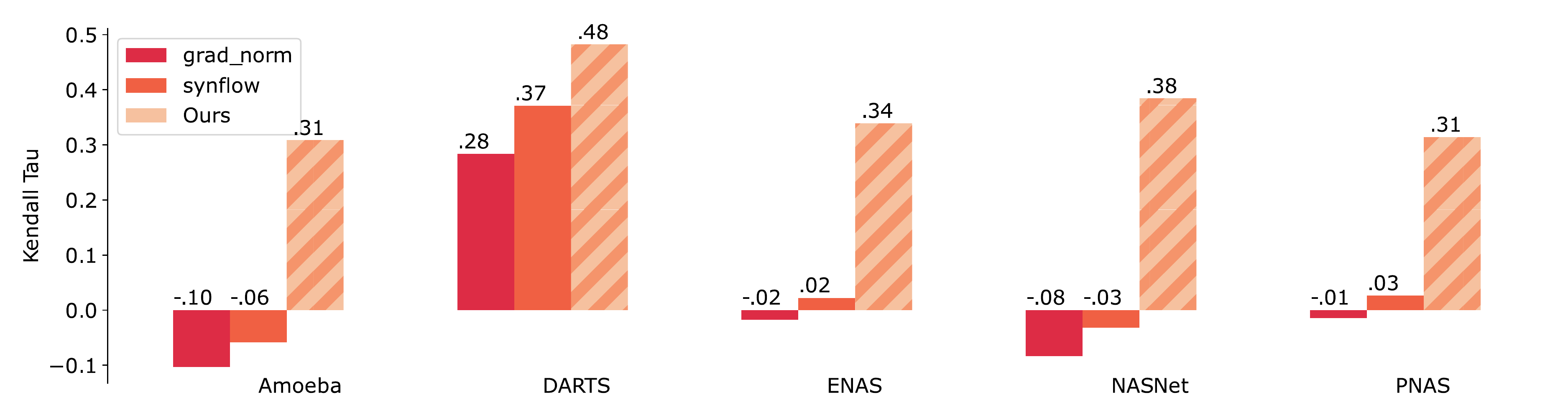}
\caption{Kendall's Tau correlation across each of the NDS CIFAR-10 search spaces. We compare our method to two alternative measures: \texttt{grad\_norm} and \texttt{synflow}. The results for \texttt{grad\_norm} refer to the absolute Euclidean-norm of the gradients over one random mini-batch of data. \texttt{synflow} is the gradient-based score defined by~\cite{tanaka2020pruning}, summed over each parameter in the network.}
\label{fig:nds_corrs}
\end{figure*}

For the search spaces across NAS-Bench-101~\citep{ying2019bench}, NAS-Bench-201~\citep{Dong2020NAS-Bench-201}, NATS-Bench SSS~\citep{dong2021nats}, and NDS~\citep{radosavovic2019network} we sample networks at random and plot our score $s$ on the {\it untrained} networks against their validation accuracies {\it when trained}. The plots for NAS-Bench-101, NAS-Bench-201, and NDS are available in Figure 3. In most cases 1000 networks are sampled.\footnote{Due to GPU memory limitations, there are 900, 749, and 973 networks shown for NDS-AmoebaNet, NDS-ENAS, and NDS-PNAS respectively.}
The plots for NATS-Bench SSS can be found in Figure~\ref{fig:extraplots-NATS} (Appendix~\ref{appendix:plots}). We also provide comparison plots for the fixed width and depth spaces in NDS in Figure~\ref{fig:extraplots-NDS} (Appendix~\ref{appendix:plots}). Kendall's Tau correlation coefficients $\tau$ are given at the top of each plot.

We find in all cases there is a positive correlation between the validation accuracy and the score. This is particularly strong for NAS-Bench-201 and NDS-DARTS. We show the Kendall's Tau correlation coefficient between $s$ and final accuracy on CIFAR-10 for NDS in Figure~\ref{fig:nds_corrs}. For comparison, we include the best-performing architecture scoring functions from~\cite{abdelfattah2021zerocost} ---~\texttt{grad\_norm} and~\texttt{synflow} --- as baselines. The first is the sum of the gradient norms for every weight in the network, and the second is the summed Synaptic Flow score derived in~\cite{tanaka2020pruning}. our score (Equation~\ref{eq:score}) correlates with accuracy across all of the search spaces, where the other two scores fluctuate substantially. These results point to our score being effective on a wide array of neural network design spaces.

\subsection{Ablation Study}
\label{sec:ablation}

\begin{figure}[!h]
    \centering
    \includegraphics[width=.9\linewidth]{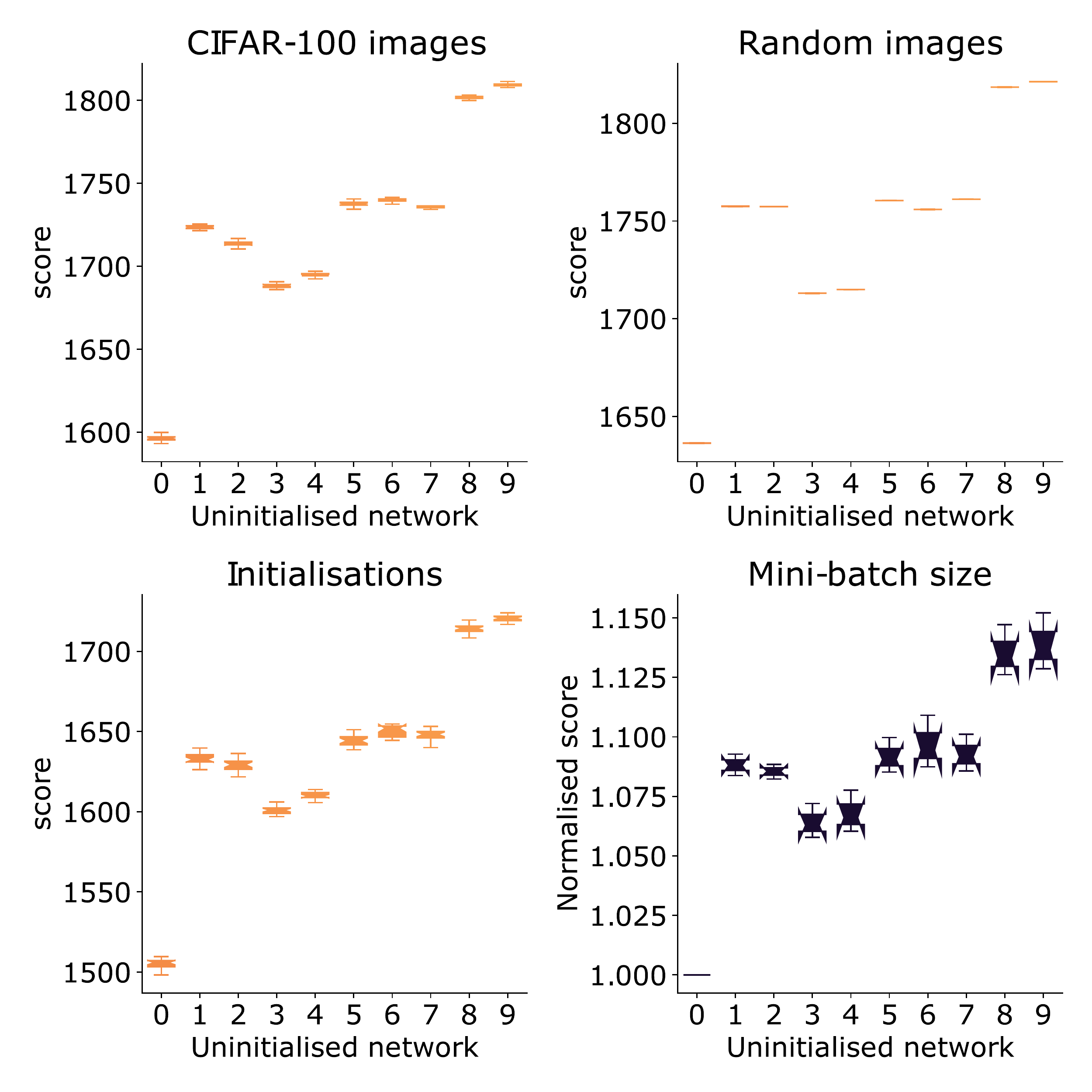}
  \caption{Ablation experiments showing the effect on our score using different CIFAR-100 mini-batches (top-left), random normally-distributed input images (top-right), weight initialisations (bottom-left), and mini-batch sizes (bottom-right) for 10 randomly selected NAS-Bench-201 architectures (one in each $5\%$ percentile range from 50-55, ..., 95-100). For each network, 20 samples were taken for each ablation. The mini-batch size was 128 for all experiments apart from the bottom-right. The bottom-right experiment used mini-batch sizes of 32, 64, 128, and 256; as the score depends on the mini-batch size we normalised the score by the minimum score of the sampled networks from the same mini-batch size.
  }
    \label{fig:ablationscore}  
    \vspace{-4mm}
\end{figure}

\textbf{How important are the images used to compute the score?\ } Since our approach relies on randomly sampling a single mini-batch of data, it is reasonable to question whether different mini-batches result in different scores. To determine whether our method is dependent on mini-batches, we randomly select 10 architectures from different CIFAR-100 accuracy percentiles in NAS-Bench-201 and compute the score separately for 20 random CIFAR-100 mini-batches. The resulting box-and-whisker plot is given in Figure~\ref{fig:ablationscore}(top-left): the ranking of the scores is reasonably robust to the specific choice of images. In Figure~\ref{fig:ablationscore}(top-right) we compute our score using normally distributed random inputs; this has little impact on the general trend. This suggests our score captures a property of the network architecture, rather than something data-specific.

\textbf{Does the score change for different initialisations?\ } Figure~\ref{fig:ablationscore}(bottom-left) shows how the score for our 10 NAS-Bench-201 architectures differs over 20 initialisations. While there is some noise, the better performing networks remain distinctive, and can be isolated.

\textbf{Does the size of the mini-batch matter?\ } As $\matr{K}_H$ scales with mini-batch size we compare across mini-batch sizes by dividing a given score by the minimum score using the same mini-batch size from the set of sampled networks.  
Figure~\ref{fig:ablationscore}(bottom-right)
presents this normalised score for different mini-batch sizes. The best performing networks remain distinct.

\begin{figure}[!h]
                \includegraphics[width=\linewidth]{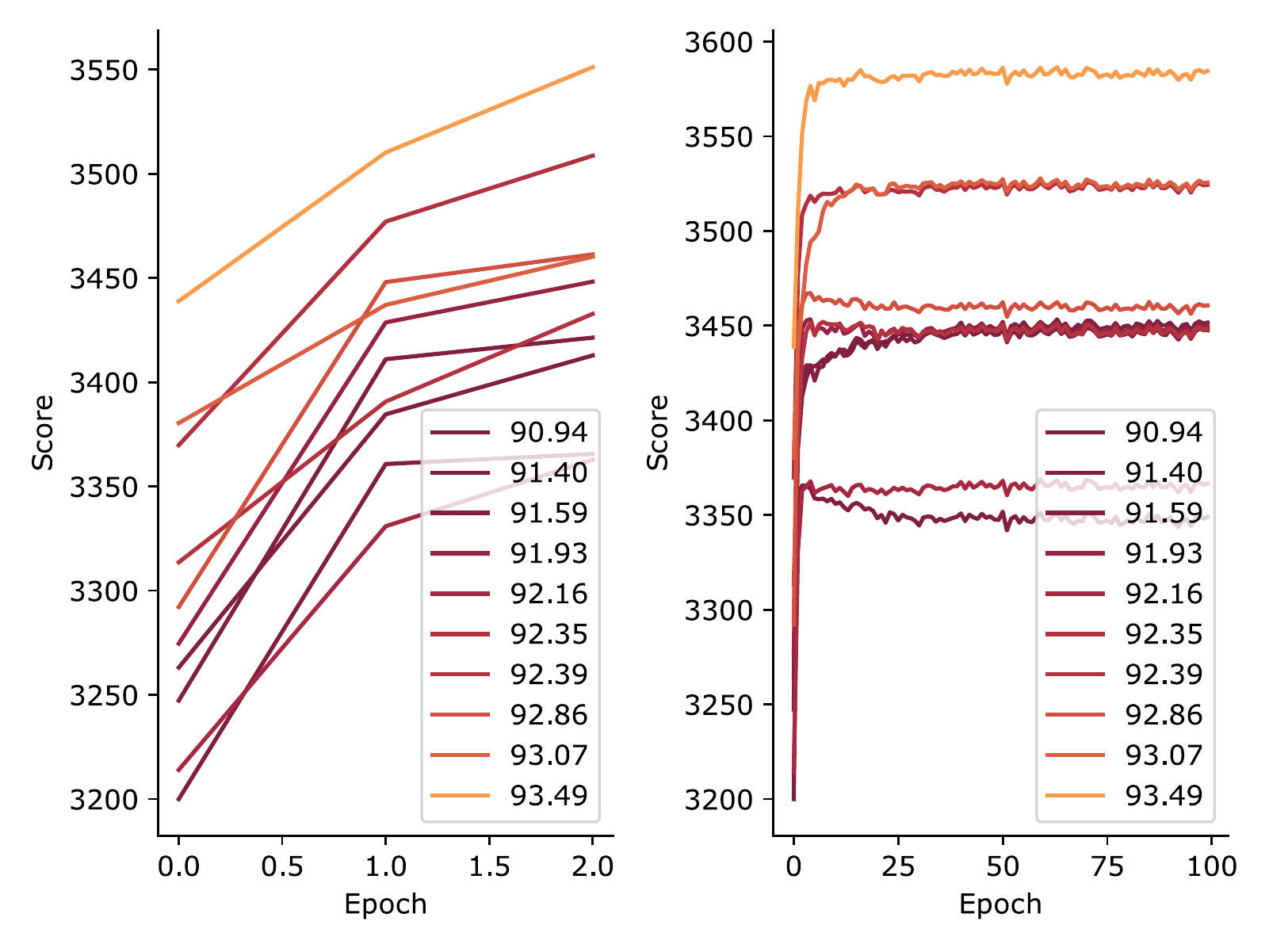}
\caption{Plots of our score (Equation~\ref{eq:score}) during training for 10 networks from NAS-Bench-201 using the CIFAR-10 dataset. The legend provides the final accuracy of the network as given by the NAS-Bench-201 API. For all 10 networks the score increases sharply in the first few epochs and then flattens. The ranking of the scores between networks remains relatively stable throughout training.}
\label{fig:scoretraj}
\end{figure}

\textbf{How does the score evolve as networks are trained?\ } Although the motivation of our work is to score uninitialised networks, it is worth observing how the score evolves as a network is trained. 
We consider 10 NAS-Bench-201 networks with $>90\%$ validation accuracy when evaluated on CIFAR-10. This makes the performance difference between the 10 networks much smaller than for the search space as a whole. 
We train each network via stochastic gradient descent with a cross entropy loss for 100 epochs and evaluate the score at each training epoch. Figure~\ref{fig:scoretraj} shows the evolution of the score. The left subplot shows a zoomed-in view of the score trajectories across the first two epochs and the right subplot shows the score trajectories across all 100 epochs. 
We observe that the score increases in all cases immediately after some training has occurred, but very quickly stabilises to a near constant value. The increase in the score value after initialisation was similar amongst the networks, and the relative ranking remained similar throughout.

In Section~\ref{sec:nas} we demonstrate how our score (Equation~\ref{eq:score}) can be used in a NAS algorithm for extremely fast search.

\section{Neural Architecture Search without Training --- NASWOT}
\label{sec:nas}

\begin{table}
\caption{Mean $\pm$ std. accuracy from NAS-Bench-101. NASWOT is our training-free algorithm (across 500 runs). REA uses evolutionary search to select an architecture (50 runs), Random selects one architecture (500 runs). AREA (assisted-REA) uses our score (Equation~\ref{eq:score}) to select the starting population for REA (50 runs).  Search times for REA and AREA were calculated using the NAS-Bench-101 API.}
\vspace{2mm}
\label{table:benchmarking101}
\centering
    \begin{tabular}{@{}llc@{}} \hline 
    Method & Search (s)  & CIFAR-10 \\
    \midrule
    Random & N/A &  90.38$\pm$5.51 \\
    NASWOT (\texttt{N}=100) & 23  & 91.77$\pm$0.05 \\
    REA      &  12000  & 93.87$\pm$0.22 \\
    AREA (Ours) &  12000  & 93.91$\pm$0.29 \\
    \midrule
    \end{tabular}
\end{table}

\begin{table*}[!h]

\caption{{\bf (a):} Mean $\pm$ std. accuracies on NAS-Bench-201. Baselines are taken directly from~\cite{Dong2020NAS-Bench-201}, averaged over 500 runs (3 for weight-sharing methods). Search times are recorded for a single 1080Ti GPU. Note that all searches are performed on CIFAR-10 before evaluating the final model on CIFAR-10, CIFAR-100, and ImageNet-16-120. The performance of our training-free approach (NASWOT) is given for different sample size~\texttt{N} (also 500 runs), along with that of our Assisted REA (AREA) approach (50 runs). We also report the results for picking a network at random, and the best possible network from the sample.~{\bf (b):} Mean $\pm$ std. accuracies (over 500 runs) on NATS-Bench SSS~\citep{dong2021nats} comparing non-weight sharing approaches to NASWOT. Unlike NAS-Bench-201, each search is performed on the same dataset that is then used to evaluate the proposed network.
}
\vspace{2mm}
\label{table:benchmarking}
\centering

\begin{adjustbox}{width=2\columnwidth,center}
\begin{tabular}{@{}llllcllcll@{}} \hline 
\multirow{2}{*}{Method} & \multirow{2}{*}{Search (s)}  & \multicolumn{2}{c}{CIFAR-10} & \phantom{ab} & \multicolumn{2}{c}{CIFAR-100} & \phantom{ab} & \multicolumn{2}{c}{ImageNet-16-120} \\
\cmidrule{3-4} \cmidrule{6-7} \cmidrule{9-10}
& & validation & test && validation & test && validation & test \\
\midrule
\multicolumn{10}{c}{\textbf{(a) NAS-Bench-201}}
\\
\midrule
\multicolumn{10}{c}{\textbf{Non-weight sharing}}\\
REA       &  12000 & 91.19$\pm$0.31 & 93.92$\pm$0.30 && 71.81$\pm$1.12 & 71.84$\pm$0.99 && 45.15$\pm$0.89 & 45.54$\pm$1.03 \\
RS        &  12000 & 90.93$\pm$0.36 & 93.70$\pm$0.36 && 70.93$\pm$1.09 & 71.04$\pm$1.07 && 44.45$\pm$1.10 & 44.57$\pm$1.25 \\
REINFORCE &  12000 & 91.09$\pm$0.37 & 93.85$\pm$0.37 && 71.61$\pm$1.12 & 71.71$\pm$1.09 && 45.05$\pm$1.02 & 45.24$\pm$1.18 \\
BOHB      &  12000 & 90.82$\pm$0.53 & 93.61$\pm$0.52 && 70.74$\pm$1.29 & 70.85$\pm$1.28 && 44.26$\pm$1.36 & 44.42$\pm$1.49 \\
\midrule

\multicolumn{10}{c}{\textbf{Weight sharing}}\\
RSPS                & 7587  & 84.16$\pm$1.69 & 87.66$\pm$1.69 && 59.00$\pm$4.60 & 58.33$\pm$4.34 && 31.56$\pm$3.28 & 31.14$\pm$3.88 \\
DARTS-V1            & 10890 & 39.77$\pm$0.00 & 54.30$\pm$0.00 && 15.03$\pm$0.00 & 15.61$\pm$0.00 && 16.43$\pm$0.00 & 16.32$\pm$0.00 \\
DARTS-V2            & 29902 & 39.77$\pm$0.00 & 54.30$\pm$0.00 && 15.03$\pm$0.00 & 15.61$\pm$0.00 && 16.43$\pm$0.00 & 16.32$\pm$0.00 \\
GDAS                & 28926 & 90.00$\pm$0.21 & 93.51$\pm$0.13 && 71.14$\pm$0.27 & 70.61$\pm$0.26 && 41.70$\pm$1.26 & 41.84$\pm$0.90 \\
SETN                & 31010 & 82.25$\pm$5.17 & 86.19$\pm$4.63 && 56.86$\pm$7.59 & 56.87$\pm$7.77 && 32.54$\pm$3.63 & 31.90$\pm$4.07 \\
ENAS                & 13315 & 39.77$\pm$0.00 & 54.30$\pm$0.00 && 15.03$\pm$0.00 & 15.61$\pm$0.00 && 16.43$\pm$0.00 & 16.32$\pm$0.00 \\
\midrule
\multicolumn{10}{c}{\textbf{Training-free}}\\

\color{Sepia}{NASWOT (\texttt{N}=10)}  &     3.05 &  89.14 $\pm$ 1.14 &  92.44 $\pm$ 1.13 &&  68.50 $\pm$ 2.03 &  68.62 $\pm$ 2.04  && 41.09 $\pm$ 3.97 &  41.31 $\pm$ 4.11  \\
\color{Sepia}{NASWOT (\texttt{N}=100)}  &     30.01 &  89.55 $\pm$ 0.89 &  92.81 $\pm$ 0.99 &&  69.35 $\pm$ 1.70 &  69.48 $\pm$ 1.70  && 42.81 $\pm$ 3.05 &  43.10 $\pm$ 3.16  \\
\color{Sepia}{NASWOT (\texttt{N}=1000)}  &     306.19 &  89.69 $\pm$ 0.73 &  92.96 $\pm$ 0.81 &&  69.86 $\pm$ 1.21 &  69.98 $\pm$ 1.22  && 43.95 $\pm$ 2.05 &  44.44 $\pm$ 2.10  \\
\midrule
Random & N/A & 83.20 $\pm$ 13.28 & 86.61 $\pm$ 13.46 && 60.70 $\pm$ 12.55 & 60.83 $\pm$ 12.58 && 33.34 $\pm$ 9.39 & 33.13 $\pm$ 9.66 \\ 
Optimal (\texttt{N}=10) & N/A & 89.92 $\pm$ 0.75 & 93.06 $\pm$ 0.59 && 69.61 $\pm$ 1.21 & 69.76 $\pm$ 1.25 && 43.11 $\pm$ 1.85 & 43.30 $\pm$ 1.87 \\
Optimal (\texttt{N}=100) & N/A & 91.05 $\pm$ 0.28 & 93.84 $\pm$ 0.23 && 71.45 $\pm$ 0.79 & 71.56 $\pm$ 0.78 && 45.37 $\pm$ 0.61 & 45.67 $\pm$ 0.64 \\ 
\midrule
\color{Sepia}{AREA}  & 12000 & 91.20 $\pm$ 0.27 & -  && 71.95 $\pm$ 0.99 &  -  && 45.70  $\pm$ 1.05 & - \\
\midrule
\multicolumn{10}{c}{\textbf{(b) NATS-Bench SSS}}
\\
\midrule
\multicolumn{10}{c}{\textbf{Non-weight sharing}}\\
REA       &  12000 & 90.37$\pm$0.20 & 93.22$\pm$0.16 && 70.23$\pm$0.50 & 70.11$\pm$0.61 && 45.30$\pm$0.69 & 45.54$\pm$0.92 \\
RS        &  12000 & 90.10$\pm$0.26 & 93.03$\pm$0.25 && 69.57$\pm$0.57 & 69.72$\pm$0.61 && 45.01$\pm$0.74 & 45.42$\pm$0.86 \\
REINFORCE &  12000 & 90.25$\pm$0.23 & 93.16$\pm$0.21 && 69.84$\pm$0.59 & 69.96$\pm$0.57 && 45.06$\pm$0.77 & 45.24$\pm$1.18 \\
BOHB      &  12000 & 90.07$\pm$0.28 & 93.01$\pm$0.24 && 69.75$\pm$0.60 & 69.90$\pm$0.60 && 45.11$\pm$0.69 & 45.56$\pm$0.81 \\
\midrule
\color{Sepia}{NASWOT (\texttt{N}=10)}  &     3.02  &  88.95 $\pm$ 0.88 &  88.66 $\pm$ 0.90 &&  64.55 $\pm$ 4.57 &  64.54 $\pm$ 4.70  && 40.22 $\pm$ 3.73 &  40.48 $\pm$ 3.73  \\
\color{Sepia}{NASWOT (\texttt{N}=100)}  &     32.36  &  89.68 $\pm$ 0.51 &  89.38 $\pm$ 0.54 &&  66.71 $\pm$ 3.05 &  66.68 $\pm$ 3.25  && 42.68 $\pm$ 2.58 &  43.11 $\pm$ 2.42  \\
\color{Sepia}{NASWOT (\texttt{N}=1000)} &          248.23 &  90.14 $\pm$ 0.30  & 93.10 $\pm$ 0.31  &&   68.96 $\pm$ 1.54 &  69.10 $\pm$ 1.61  && 44.57 $\pm$ 1.48 &  45.08 $\pm$ 1.55  \\
\midrule
\end{tabular}
\end{adjustbox}

\end{table*}

In Section~\ref{sec:scoring} we derived a score for cheaply ranking networks at initialisation based on their expected performance (Equation~\ref{eq:score}). Here as a proof of concept, we integrate this score into a simple search algorithm and evaluate its ability to alleviate the need for training in NAS. Code for reproducing our experiments is available at~\url{https://github.com/BayesWatch/nas-without-training}.

Many NAS algorithms are based on that of~\cite{zoph2017neural}; it uses a generator network which proposes architectures. The weights of the generator are learnt by training the networks it generates, either on a proxy task or on the dataset itself, and using their trained accuracies as signal through e.g.\ REINFORCE~\citep{williams1992simple}. This is repeated until the generator is trained; it then produces a final network which is the output of this algorithm. The vast majority of the cost is incurred by having to train candidate architectures for every single controller update. Note that there exist alternative schema utilising e.g.\ evolutionary algorithms~\citep{real2019regularized} or bilevel optimisation~\citep{liu2019darts} but all involve training.

We instead propose a simple alternative---NASWOT---illustrated in Algorithm~\ref{algo:ours}. Instead of having a neural network as a generator, we randomly propose a candidate from the search space and then rather than training it, we score it in its untrained state using Equation~\ref{eq:score}. We do this \texttt{N} times---i.e.\ we have a sample size of \texttt{N} architectures---and then output the highest scoring network.

\begin{figure}
{\tiny
\vspace{-4mm}
    \begin{algorithm}[H]
    \caption{NASWOT}
    \begin{algorithmic}[h]
          \State  generator = RandomGenerator() 
                \State best\_net, best\_score = None, 0

          \For{\texttt{i=1:N}}
          
              \State net = generator.generate()
        \State \textcolor{teal}{score = net.score()} 
        \If{score $>$ best\_score}
         \State best\_net, best\_score = net, score 
        \EndIf
        \EndFor
    \State chosen\_net = best\_network

    \end{algorithmic}
    
    \label{algo:ours}
    \end{algorithm}
    
    }
    \vspace{-8mm}
\end{figure}

\textbf{NAS-Bench-101.\ } We compare NASWOT to 12000 seconds of REA \citep{real2019regularized} and random selection on NAS-Bench-101~\citep{ying2019bench} in Table~\ref{table:benchmarking101}. NASWOT can find a network with a final accuracy roughly midway between these methods in under a minute on a single GPU.

\textbf{NAS-Bench-201.\ }~\citet{Dong2020NAS-Bench-201} benchmark a wide range of NAS algorithms, both with and without weight sharing, that we compare to NASWOT. The weight sharing methods are random search with parameter sharing (RSPS,~\citealp{li2019random}), first-order DARTS (DARTS-V1,~\citealp{liu2019darts}), second order DARTS (DARTS-V2,~\citealp{liu2019darts}), GDAS~\citep{dong2019searching}, SETN~\citep{dong2019one}, and ENAS~\citep{pham2018efficient}. The non-weight sharing methods are random search with training (RS), REA~\citep{real2019regularized}, REINFORCE~\citep{williams1992simple}, and BOHB~\citep{falkner2018bohb}. For implementation details we refer the reader to~\cite{Dong2020NAS-Bench-201}. The hyperparameters in NAS-Bench-201 are fixed --- these results may not be invariant to hyperparameter choices, which may explain the low performance of e.g.\ DARTS.

All searches are performed on CIFAR-10, and the output architecture is then trained and evaluated on each of CIFAR-10, CIFAR-100, and ImageNet-16-120 for different dataset splits. We report results in Table~\ref{table:benchmarking}(a). Search times are reported for a single GeForce GTX 1080 Ti GPU. As per the NAS-Bench-201 setup, the non-weight sharing methods are given a time budget of 12000 seconds. For NASWOT and the non-weight sharing methods, accuracies are averaged over 500 runs. For weight-sharing methods, accuracies are reported over 3 runs. We report NASWOT for sample sizes of \texttt{N}=10, \texttt{N}=100, and  \texttt{N}=1000. NASWOT is able to outperform all of the weight sharing methods while requiring a fraction of the search time.

The non-weight sharing methods do outperform NASWOT, though they also incur a large search time cost. It is encouraging however, that in a matter of seconds, NASWOT is able to find networks with performance close to the best non-weight sharing algorithms, suggesting that network architectures themselves contain almost as much information about final performance at initialisation as after training.

Table~\ref{table:benchmarking}(a) also shows the effect of sample size (\texttt{N}). We show the accuracy of networks chosen by our method for each~\texttt{N}. We list optimal accuracy for each~\texttt{N}, and random selection over the whole benchmark, both averaged over 500 runs. We observe that sample size increases NASWOT performance.

\textbf{NATS-Bench.\ }~\citet{dong2021nats} expanded on the original NAS-Bench-201 work to include a further search space: NATS-Bench SSS. The same non-weight sharing methods were evaluated as in NAS-Bench-201. Unlike NAS-Bench-201, search and evaluation are performed on the same dataset. Implementation details are available in~\cite{dong2021nats}. 
We observe that sample size increases NASWOT performance significantly on NATS-Bench SSS, to the point where it is extremely similar to other methods.

A key practical benefit of NASWOT is its rapid execution time. This may be important when repeating NAS several times, for instance for several hardware devices or datasets. This affords us the ability in future to specialise neural architectures for a task and resource environment cheaply, demanding only a few seconds per setup. Figure~\ref{fig:timeacc} shows our method in contrast to other NAS methods for NAS-Bench-201, showing the trade-off between final network accuracy and search time.

\subsection{Assisted Regularised EA --- AREA}
\label{sec:area}

Our proposed score can be straightforwardly incorporated into existing NAS algorithms. To demonstrate this we implemented a variant of REA~\citep{real2019regularized}, which we call Assisted-REA (AREA). REA starts with a randomly-selected population (10 in our experiments). AREA instead randomly-samples a larger population (in our experiments we double the randomly-selected population size to 20) and uses our score (Equation~\ref{eq:score}) to select the initial population (of size 10) for the REA algorithm. Pseudocode can be found in Algorithm~\ref{algo:oursrea} with results on NAS-Bench-101 and NAS-Bench-201 in Tables~\ref{table:benchmarking101} and~\ref{table:benchmarking}. AREA outerforms REA on NAS-Bench-201 (CIFAR-100, ImageNet-16-120) but is very similar to REA on NAS-Bench-101. We hope that future work will build on this algorithm further.

\begin{figure}
\vspace{-2mm}
{\tiny
      \begin{algorithm}[H]
    \caption{Assisted Regularised EA --- AREA}
    \begin{algorithmic}[h]
           \State  population = []
                 \State  generator = RandomGenerator()
                 \For{\texttt{i=1:M}} 
              \State net = generator.generate() 
              \State \textcolor{teal}{scored\_net = net.score()}
              \State population.append(scored\_net)
          \EndFor
          \State Keep the top N scored networks in the population
          \State history = []
          \For{\texttt{net in population}} 
              
              \State \textcolor{purple}{trained\_net = net.train()}
              \State history.append(trained\_net)
          \EndFor
          \While{time limit not exceeded}
               
               \State Sample sub-population, S, without replacement from population
               \State Select network in S with highest accuracy as parent
               \State Mutate parent network to produce child
               \State Train child network
               \State Remove oldest network from population 
               \State population.append(child network)
               \State history.append(child network)
               
          \EndWhile
    \State chosen\_net = Network in history with highest accuracy
    \end{algorithmic}
    
    \label{algo:oursrea}
    \end{algorithm}
}
\vspace{-3mm}
\end{figure}

\begin{figure}[!h]
\centering
                \includegraphics[width=.9\linewidth]{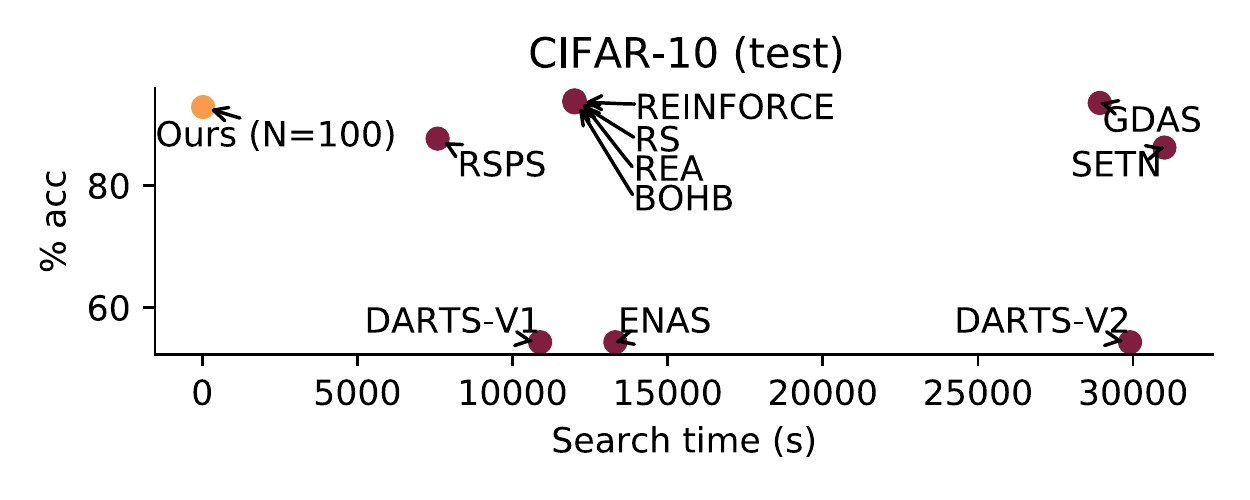}
\vspace{-3mm}
\caption{Plot showing the search time (as measured using a 1080Ti) against final accuracy of the proposed NAS-Bench-201 network for a number of search strategies.}
\label{fig:timeacc}
\vspace{-5mm}
\end{figure}

\vspace{-3mm}
\section{Conclusion}

NAS has previously suffered from intractable search spaces and heavy search costs. Recent advances in producing tractable search spaces, through NAS benchmarks, have allowed us to investigate if such search costs can be avoided. In this work, we have shown that it is possible to navigate these spaces with a search algorithm---NASWOT---in a matter of seconds, relying on simple, intuitive observations made on initialised neural networks, that challenges more expensive black box methods involving training. Future applications of this approach to architecture search may allow us to use NAS to specialise architectures over multiple tasks and devices without the need for long training stages.
We also demonstrate how our approach can be combined into an existing NAS algorithm. This work is not without its limitations; our scope is restricted to convolutional architectures for image classification. However, we hope that this will be a powerful first step towards removing training from NAS and making architecture search cheaper, and more readily available to practitioners.

\vspace{-2mm}
\paragraph{Acknowledgements.} { This work was supported in part by the EPSRC Centre for Doctoral Training in Pervasive Parallelism and a Huawei DDMPLab Innovation Research Grant. The authors are grateful to Eleanor Platt, Massimiliano Patacchiola, Paul Micaelli, and the anonymous reviewers for their helpful comments. The authors would like to thank Xuanyi Dong for correspondence on NAS-Bench-201.}

\bibliography{main}
\bibliographystyle{icml2021}
\clearpage 
\appendix

\section{NAS Benchmarks}
\label{appendix:nasbench}

\subsection{NAS-Bench-101}
In NAS-Bench-101, the search space is restricted as follows: algorithms must search for an individual cell which will be repeatedly stacked into a pre-defined skeleton, shown in Figure~\ref{fig:skelegrow}. Each cell can be represented as a directed acyclic graph (DAG) with up to 9 nodes and up to 7 edges. Each node represents an operation, and each edge represents a state. Operations can be chosen from: $3\times 3$ convolution, $1\times 1$ convolution, $3\times 3$ max pool. An example of this is shown in Figure~\ref{fig:nasdag101}. After de-duplication, this search space contains 423,624 possible neural networks. These have been trained exhaustively, with three different initialisations, on the CIFAR-10 dataset for 108 epochs. 

\subsection{NAS-Bench-201}
\label{sec:nasbench201}

\begin{figure}[!h]
        \begin{subfigure}[b]{0.5\textwidth}
        \centering
         \includegraphics[width=.5\linewidth]{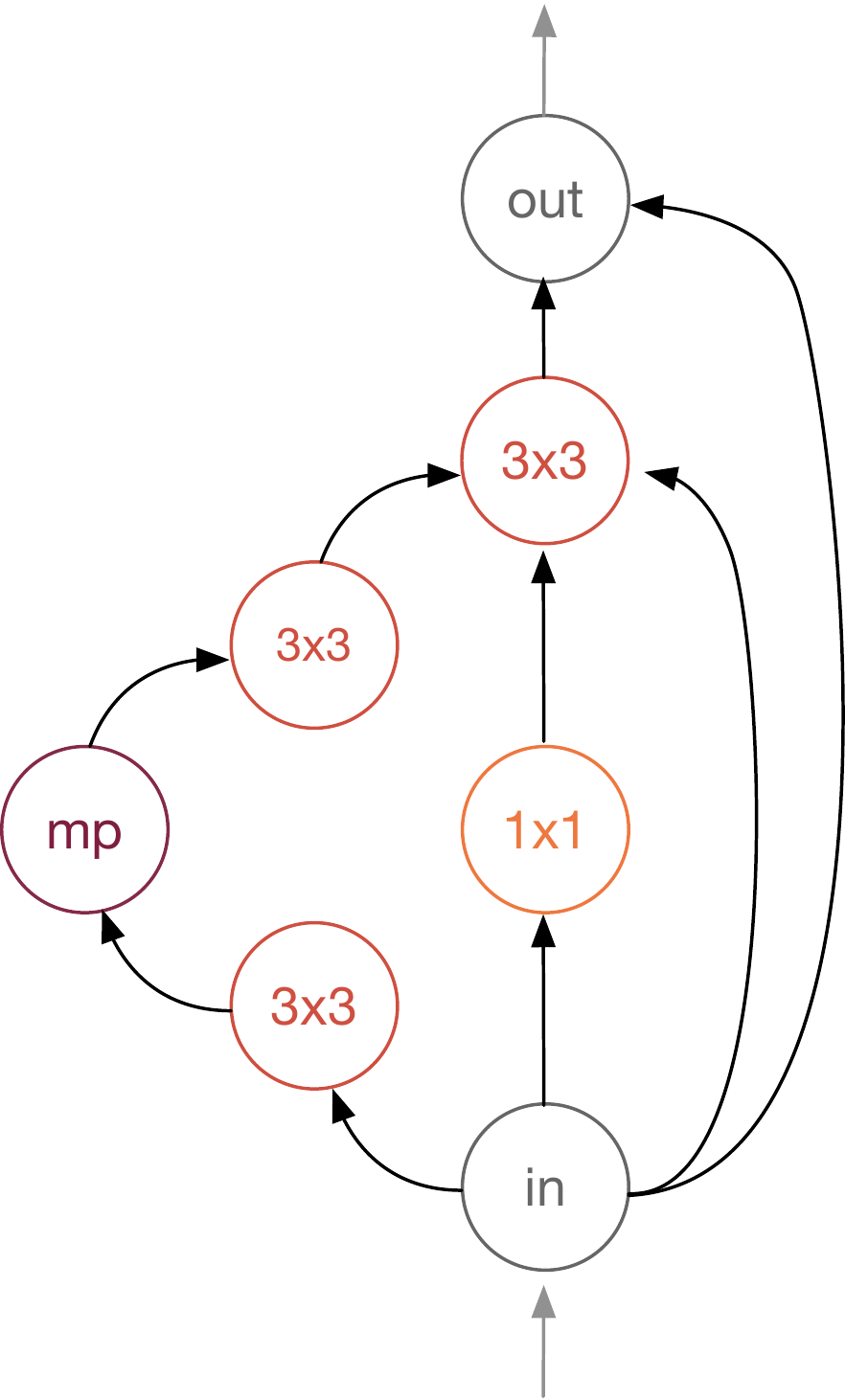}
            \subcaption{A NAS-Bench-101 cell.}\label{fig:nasdag101}
                \includegraphics[width=\linewidth]{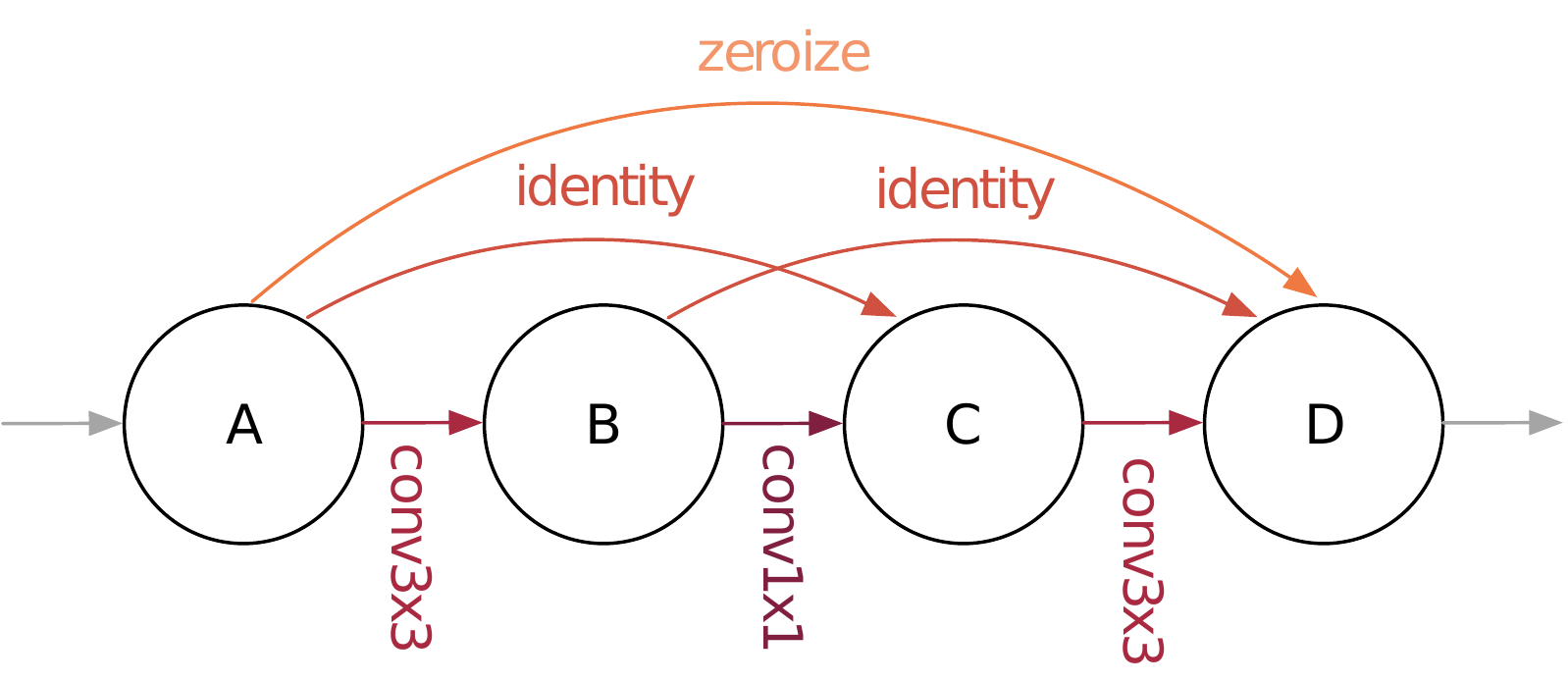}
                \subcaption{A NAS-Bench-201 and NATS-Bench TSS cell.}\label{fig:nasdag201}
                \includegraphics[width=.9\linewidth,right]{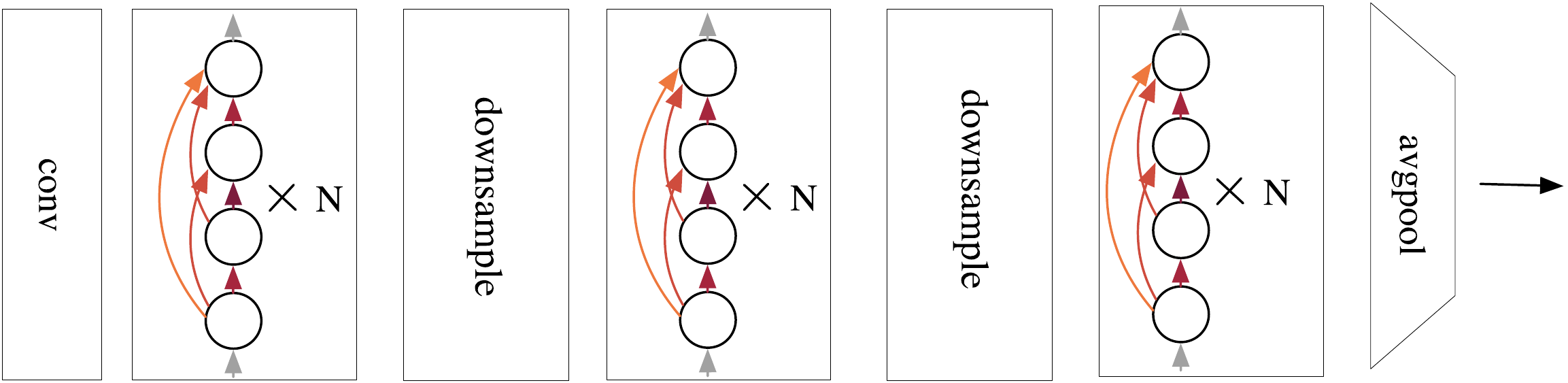}
                \subcaption{The skeleton for NAS-Bench-101 (N=3), 201 (N=5), and NATS-Bench TSS (N=5).}\label{fig:skelegrow}
       \end{subfigure}

\caption{(a): An example cell from NAS-Bench-101, represented as a directed acyclic graph. The cell has an input node, an output node, and 5 intermediate nodes, each representing an operation and connected by edges. Cells can have at most 9 nodes and at most 7 edges. NAS-Bench-101 contains 426k possible cells. By contrast, (b) shows a NAS-Bench-201 (NATS-Bench TSS) cell, which uses nodes as intermediate states and edges as operations. The cell consists of an input node (A), two intermediate nodes (B, C) and an output node (D). An edge e.g.\ A$\rightarrow$ B performs an operation on the state at A and adds it to the state at B. Note that there are 6 edges, and 5 possible operations allowed for each of these. This gives a total of $5^6$ or 15,625 possible cells. (c): Each cell is the constituent building block in an otherwise-fixed network skeleton (where N=5). As such, NAS-Bench-201 contains 15,625 architectures.}
\label{fig:test}
\end{figure}

In NAS-Bench-201, networks also share a common skeleton (Figure~\ref{fig:skelegrow}) that consists of stacks of its unique {\it cell} interleaved with fixed residual downsampling blocks. Each cell (Figure~\ref{fig:nasdag201}) can be represented as a densely-connected DAG of 4 ordered nodes (A, B, C, D) where node A is the input and node D is the output. In this graph, there is an edge connecting each node to all subsequent nodes for a total of 6 edges, and each edge can perform one of 5 possible operations (Zeroise, Identity, $3\times 3$ convolution, $1\times 1$ convolution, $3\times 3$ average pool). The search space consists of every possible cell. As there are 6 edges, on which there may be one of 5 operations, this means that there are $5^6 = 15,625$ possible cells. This makes for a total of 15,625 networks as each network uses just one of these cells repeatedly. The authors have manually split CIFAR-10, CIFAR-100, and ImageNet-16-120~\citep{chrabaszcz2017downsampled} into train/val/test, and provide full training results across all networks for (i) training on train, evaluation on val, and (ii) training on train/val, evaluation on test. The split sizes are 25k/25k/10k for CIFAR-10, 50k/5k/5k for CIFAR-100, and 151.7k/3k/3k for ImageNet-16-120.

\subsection{NATS-Bench}
\label{sec:natsbench}
NATS-Bench~\citep{dong2021nats} comprises two search spaces: a topology search space and a size search space. The networks in both spaces share a common skeleton which is the same as the skeleton used in NAS-Bench-201. The topology search space (NATS-Bench TSS) is the same as NAS-Bench-201 whereby networks vary by operation comprising the network cell. 
The size search space instead varies the channels of layers in 5 blocks of the skeleton architecture. Every network uses the same cell operations. The choice of operations corresponds to the best performing network in the topology search space with respect to the CIFAR-100 dataset. For each block the layer channel size is chosen from 8 possible sizes (8, 16, 24, 32, 40, 48, 56, 64). This leads to $8^5 = 32768$ networks in the size search space (NATS-Bench SSS).

\clearpage
\onecolumn
\section{Additional Plots}
\label{appendix:plots}

\begin{figure}[!h]

    \centering
    \begin{subfigure}[b]{0.3\textwidth}
        \includegraphics[width=\textwidth]{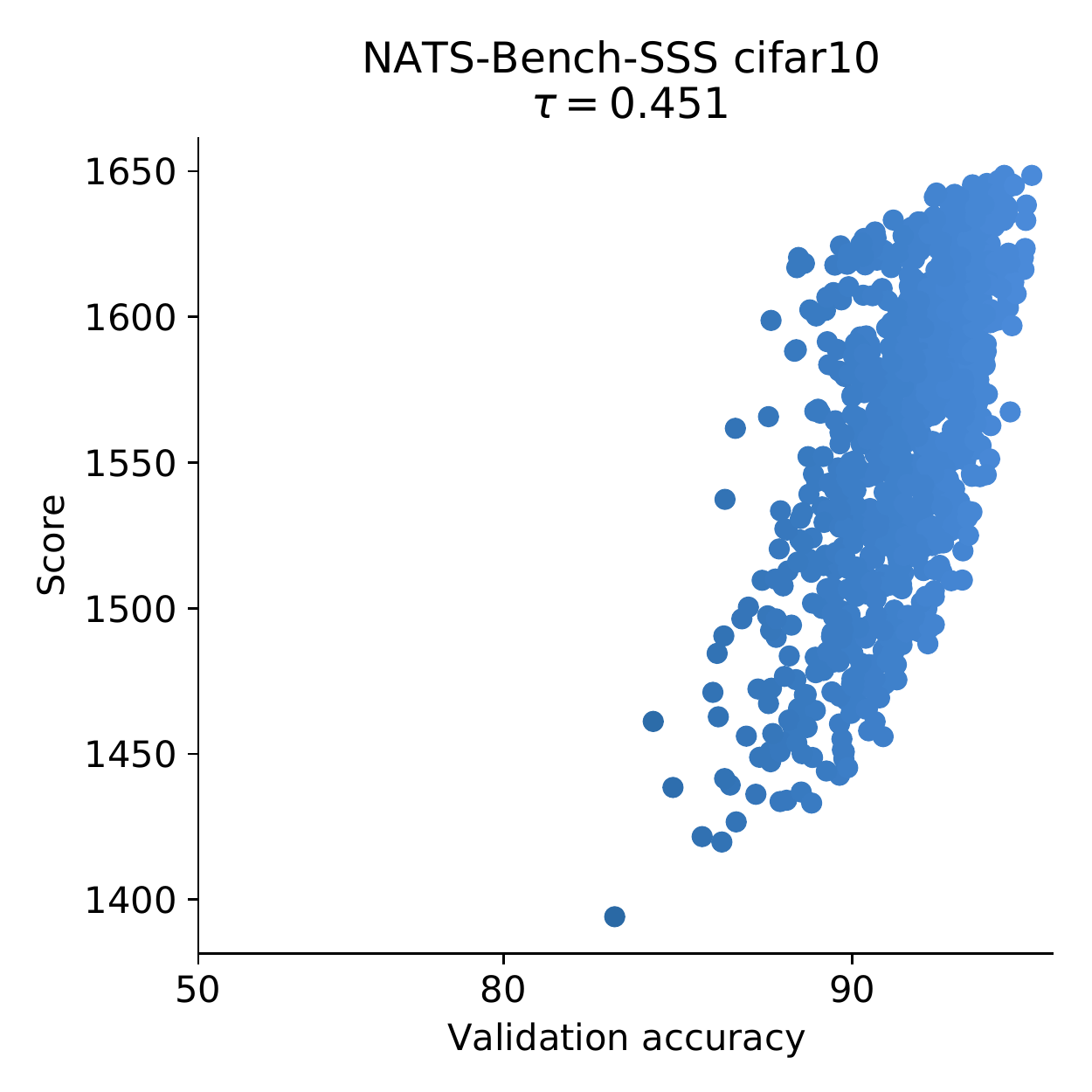}
        \caption{}
    \end{subfigure}
    \hfill
    \begin{subfigure}[b]{0.3\textwidth}
         \centering
         \includegraphics[width=\textwidth]{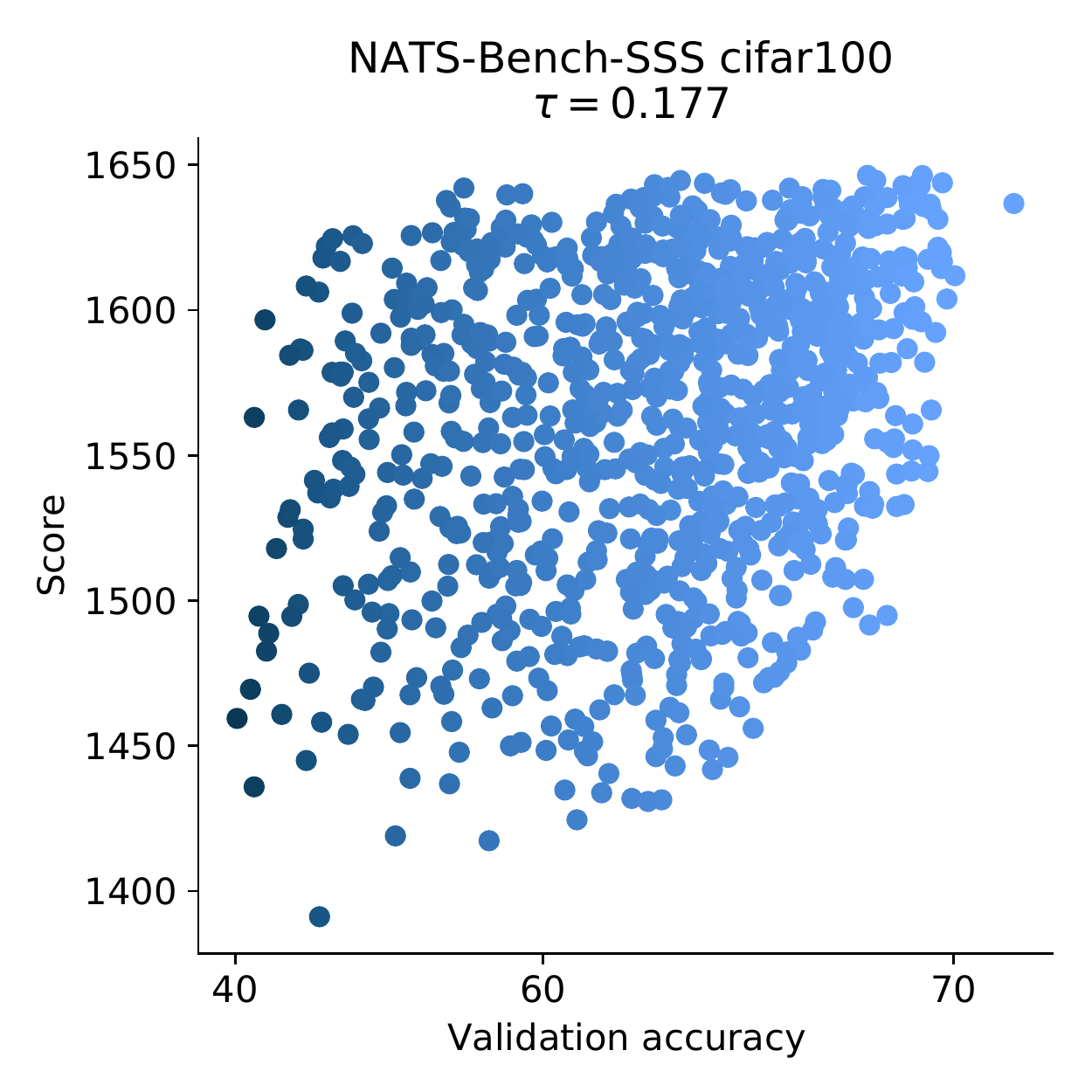}
         \caption{}
     \end{subfigure}
     \hfill
     \begin{subfigure}[b]{0.3\textwidth}
         \centering
         \includegraphics[width=\textwidth]{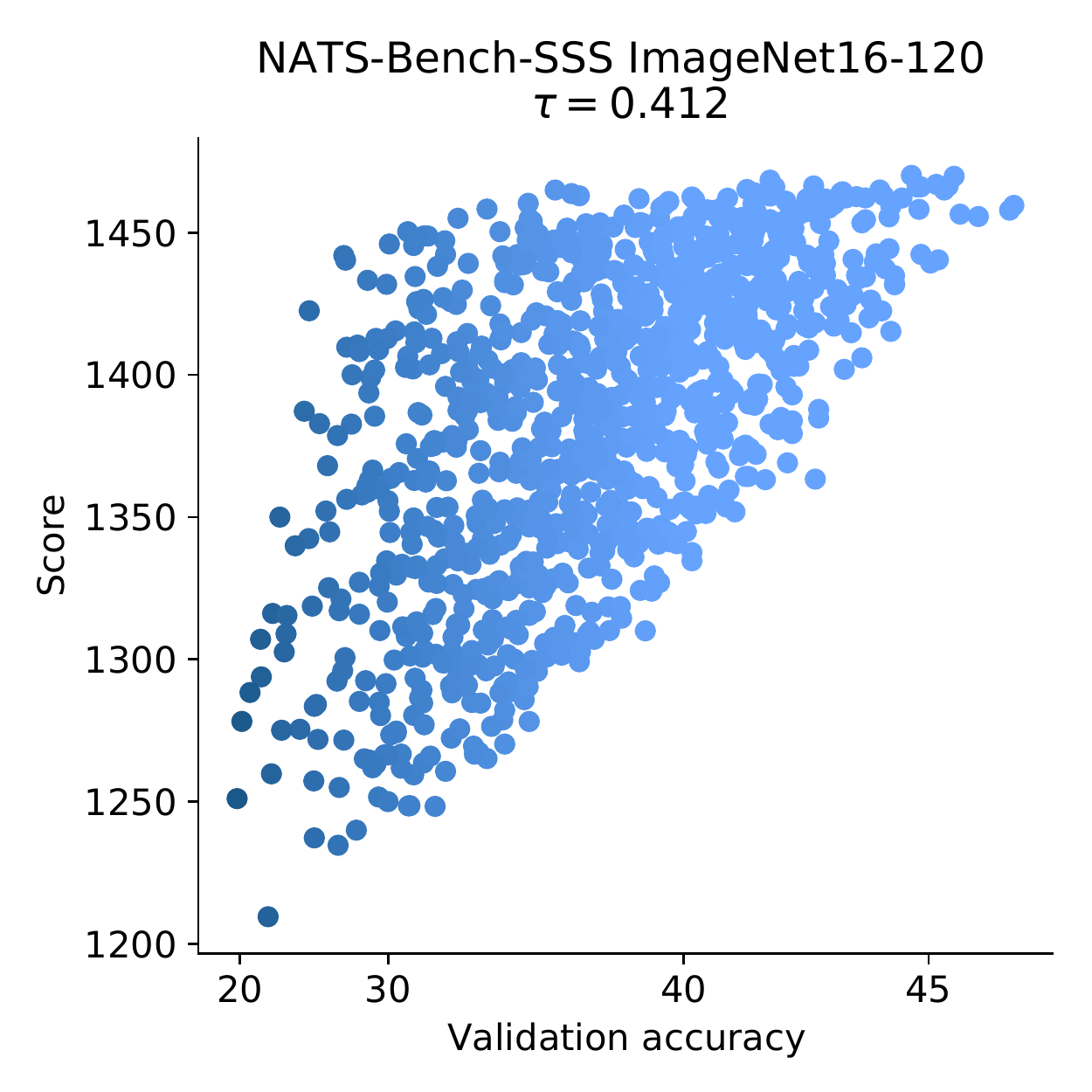}
         \caption{}
     \end{subfigure}
    
    \caption{Further plots of our score (Equation~\ref{eq:score}) for 1000 randomly sampled {\bf untrained} architectures in 
   NATS-Bench SSS
   against validation accuracy when trained on (a) CIFAR-10, (b) CIFAR-100, and (c) ImageNet16-120.}
    \label{fig:extraplots-NATS} 
\end{figure}

\vspace{-5cm}

\begin{figure}[!b]
\vspace{-2cm}

    \centering
     \begin{subfigure}[b]{0.3\textwidth}
         \centering
         \includegraphics[width=\textwidth]{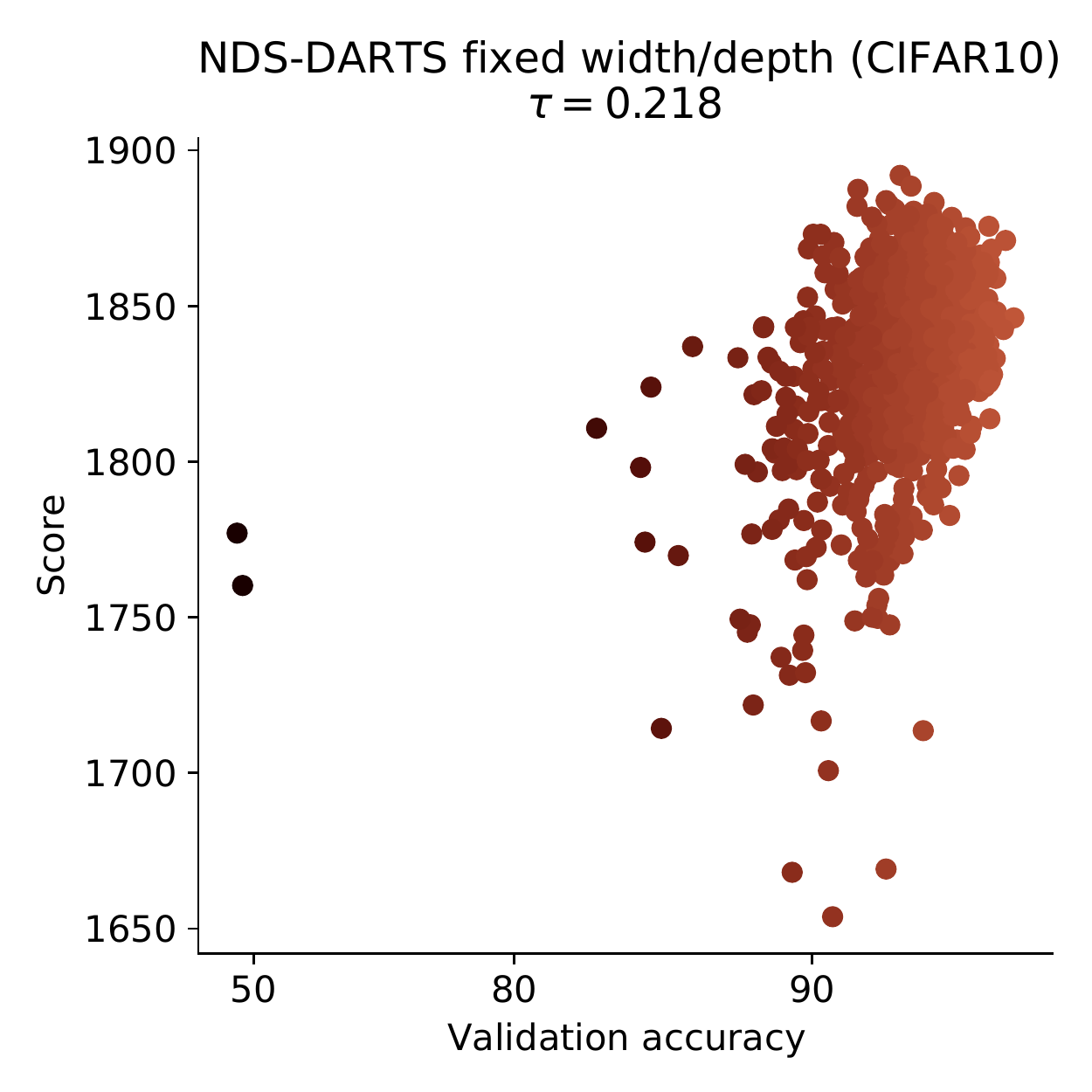}
         \caption{}
     \end{subfigure}
     \hfill
          \begin{subfigure}[b]{0.3\textwidth}
         \centering
         \includegraphics[width=\textwidth]{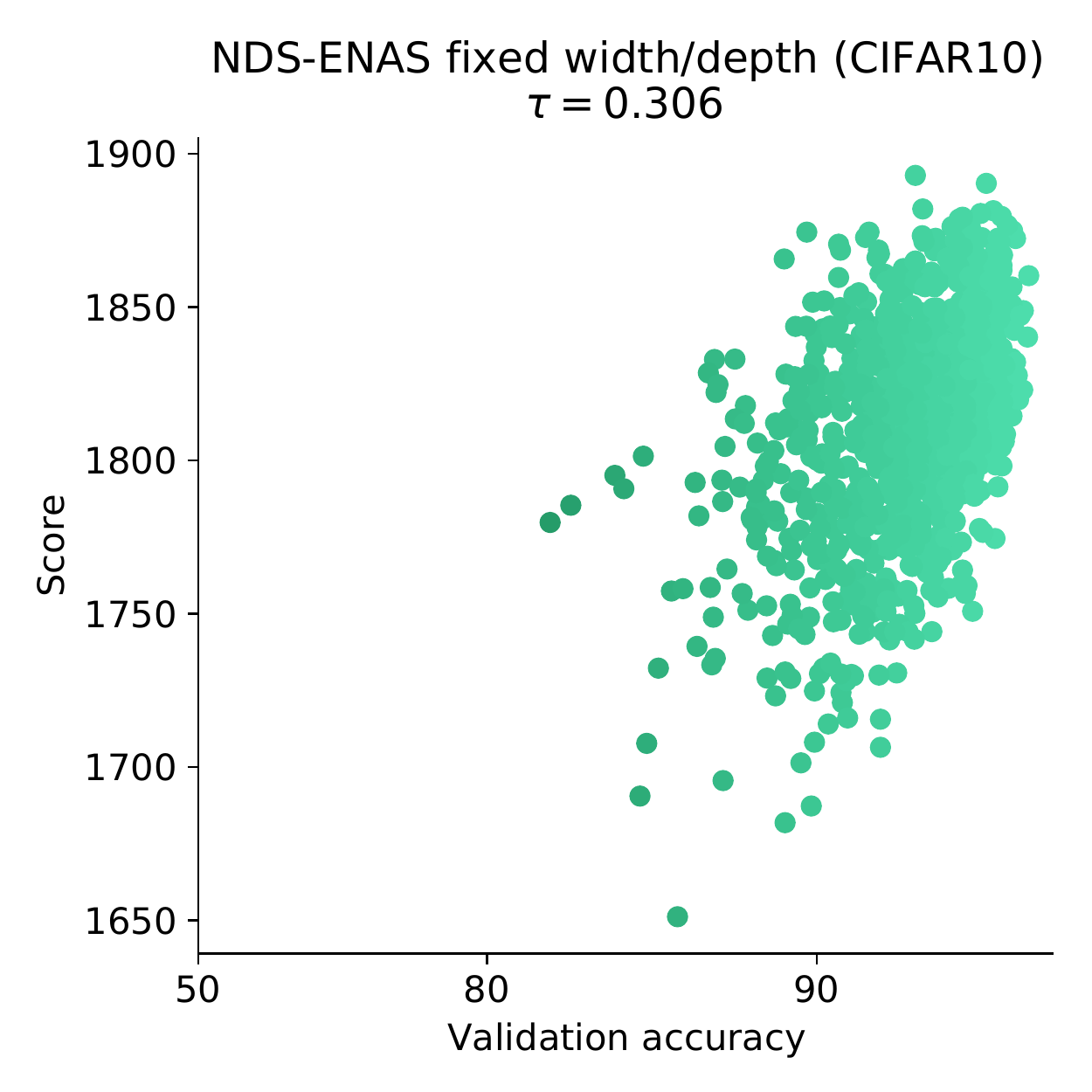}
         \caption{}
     \end{subfigure}
     \hfill
     \begin{subfigure}[b]{0.3\textwidth}
         \centering

         \includegraphics[width=\textwidth]{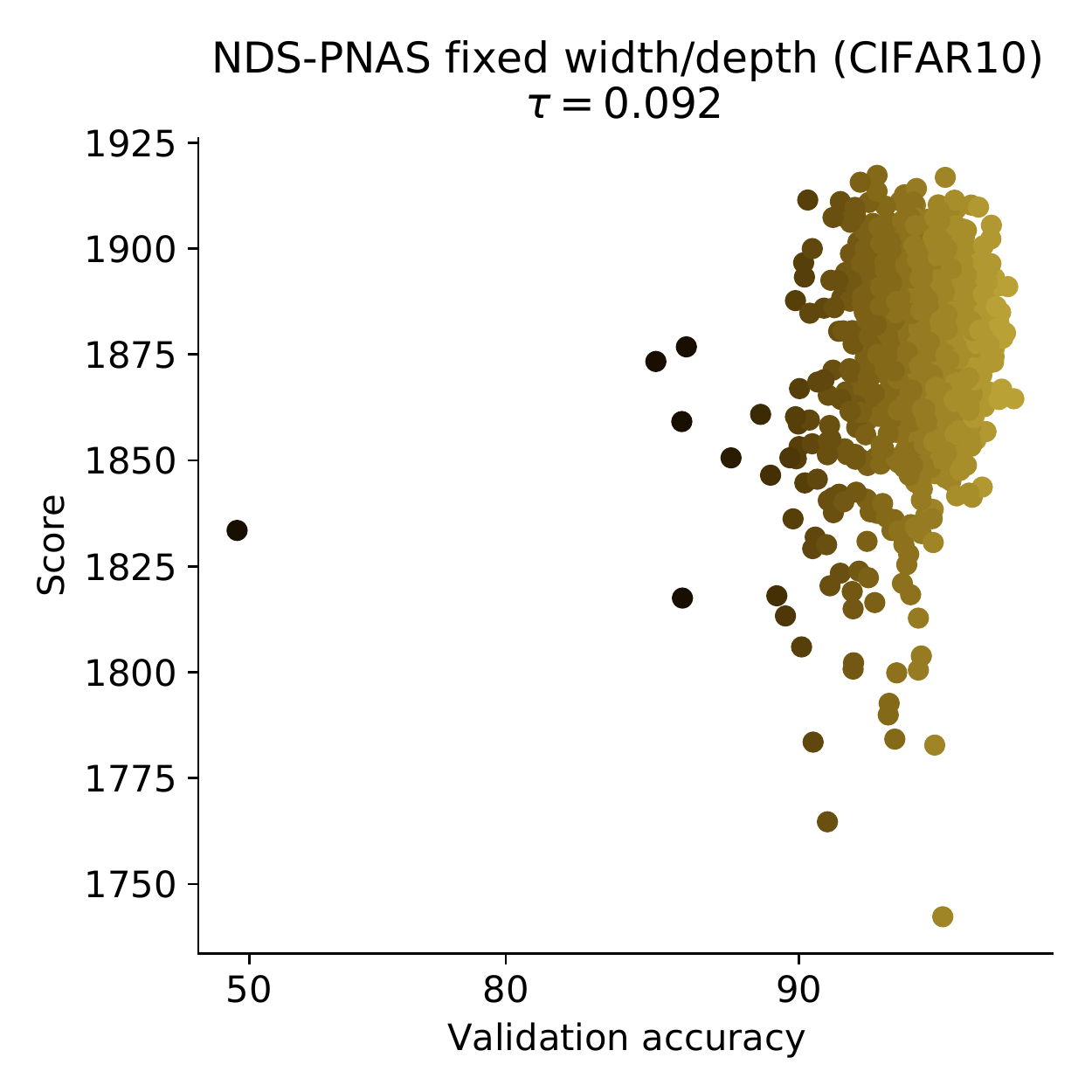}
         \caption{}
     \end{subfigure}
    
     \hfill
     \vspace{-.5cm}

    \centering
     \begin{subfigure}[b]{0.3\textwidth}
         \centering
         \includegraphics[width=\textwidth]{figures/naswot_hook_logdet_nds_darts_cifar10_none_0.05_1_True_128_1_1-min.pdf}
         \caption{}
     \end{subfigure}
     \hfill
          \begin{subfigure}[b]{0.3\textwidth}
         \centering
         \includegraphics[width=\textwidth]{figures/naswot_hook_logdet_nds_enas_cifar10_none_0.05_1_True_128_1_1-min.pdf}
         \caption{}
     \end{subfigure}
     \hfill
     \begin{subfigure}[b]{0.3\textwidth}
         \centering
         \includegraphics[width=\textwidth]{figures/naswot_hook_logdet_nds_pnas_cifar10_none_0.05_1_True_128_1_1-min.pdf}
         \caption{}
     \end{subfigure}
    
     \hfill
     
     \caption{Further plots of our score (Equation~\ref{eq:score}) for around 1000 randomly sampled {\bf untrained} architectures in 
   NDS-DARTS,
   NDS-ENAS, and
   NDS-PNAS 
   against validation accuracy when trained. The top row shows the fixed width and depth variants of the search spaces, while the bottom row shows the variable width and depth spaces.}
    \label{fig:extraplots-NDS} 
\end{figure}

\end{document}